\documentclass{ecai} 

\usepackage{latexsym}
\usepackage{amssymb}
\usepackage{amsmath}
\usepackage{amsthm}
\usepackage{booktabs}
\usepackage{enumitem}
\usepackage{graphicx}
\usepackage{color}

\usepackage{multicol}
\usepackage{multirow,tabularx}
\usepackage{comment}
\usepackage[dvipsnames]{xcolor}

\def\onedot{.}
\def\eg{\emph{e.g}\onedot}

\def\etal{\emph{et al}\onedot}

\newcommand{\btextbf}[1]{{\color{blue}\textbf{#1}}}

\newcommand{\BibTeX}{B\kern-.05em{\sc i\kern-.025em b}\kern-.08em\TeX}

\begin{document}


\begin{frontmatter}


\paperid{m577} 


\title{Advancing Wildfire Risk Prediction via Morphology-Aware Curriculum Contrastive Learning}


\author[A]{\fnms{Fabrizio}~\snm{{Lo~Scudo}}\thanks{Corresponding Authors: \{fabrizio.loscudo, alessio.derango\}@unical.it}}
\author[A]{\fnms{Alessio}~\snm{De Rango}}
\author[A]{\fnms{Luca}~\snm{Furnari}}
\author[A]{\fnms{Alfonso}~\snm{Senatore}}
\author[A]{\fnms{Donato}~\snm{D'Ambrosio}}
\author[A]{\fnms{Giuseppe}~\snm{Mendicino}}
\author[A]{\fnms{Gianluigi}~\snm{Greco}}

\address[A]{University of Calabria}


\begin{abstract}
Wildfires significantly impact natural ecosystems and human health, leading to biodiversity loss, increased hydrogeological risks, and elevated emissions of toxic substances. 
Climate change exacerbates these effects, particularly in regions with rising temperatures and prolonged dry periods, such as the Mediterranean. 
This requires the development of advanced risk management strategies that utilize state-of-the-art technologies.
However, in this context, the data show a bias toward an imbalanced setting, where the incidence of wildfire events is significantly lower than typical situations. This imbalance, coupled with the inherent complexity of high-dimensional spatio-temporal data, poses significant challenges for training deep learning architectures.
Moreover, since precise wildfire predictions depend mainly on weather data, finding a way to reduce computational costs to enable more frequent updates using the latest weather forecasts would be beneficial.
This paper investigates how adopting a contrastive framework can address these challenges through enhanced latent representations for the patch's dynamic features. 
We thus introduce a new morphology-based curriculum contrastive learning that mitigates issues associated with diverse regional characteristics and enables the use of smaller patch sizes without compromising performance.
An experimental analysis is performed to validate the effectiveness of the proposed modeling strategies.
\end{abstract}

\end{frontmatter}


\section{Introduction}
\label{sec:intro}

Wildfires have a tremendous impact on natural ecosystems and human health.  They induce a loss of biodiversity, due to the destruction of plants, animals, and soil~\citep{Pellegrini2018}; they lead to an increase of hydrogeological risks, because of soil impermeabilization and reduced slope stability~\citep{Shakesby2011}; and they cause elevated emissions of toxic substances that are harmful to humans~\citep{Reid2016}. Climate change increases the probability of wildfires in regions where rising temperatures are paired with extended dry periods, such as the Mediterranean area~\citep{Mendicino20071409,SENATORE2022101120}; indeed, dry vegetation acts as fuel for wildfires, aiding their ignition and spread~\citep{Climate2023}. In this context,  it is crucial to devise novel and more effective risk management strategies, by taking advantage of state-of-the-art technologies. In fact, the rise in average temperatures and the increase in the duration of hot seasons could reduce the effectiveness of existing wildfire protection programs and activities~\citep{Faivre2018}.

Enhancing wildfire risk management requires accurate forecasting of the likelihood and the spread of wildfires. 
The most renowned and widely used one is the \emph{Canadian Fire Weather Index (FWI)}~\citep{van1987development}, which has been adopted since 2007 within the European Forest Fire Information System (EFFIS) network~\citep{EFFIS}.
The Canadian Index exploits two kinds of data, respectively populated by ``dynamic'' and ``static'' variables.
The former category includes, for instance, wind speed/direction and air temperature, relative humidity, and precipitation: these variables change over time and are usually collected on an hourly scale and then summarized on a daily scale.
Instead, static variables report information on the morphology of the territory of interest, such as the elevation above sea 
level, the slope, the land exposure, and - hence - they are assumed to be constant in time.
FWI-based forecasting systems, implemented at both continental and local scales, have demonstrated significant effectiveness in real-world applications. Recent research has focused on enhancing these systems’ quality and reliability, particularly by developing complex models that consider additional triggering factors. Consequently, deep learning models have emerged to predict wildfire risk indices~\citep{de2023application}, incorporating anthropic variables such as proximity to urban centers and roads.
The most noticeable examples are the convolutional \emph{LSTM}-based approach of Kondylatos \etal~\citep{Kondylatos2022}, and \emph{2D-3D CNN} framework of Eddin \etal~\citep{eddin2023location}.

\begin{figure}[ht]
\centering
\includegraphics[width=8.cm]{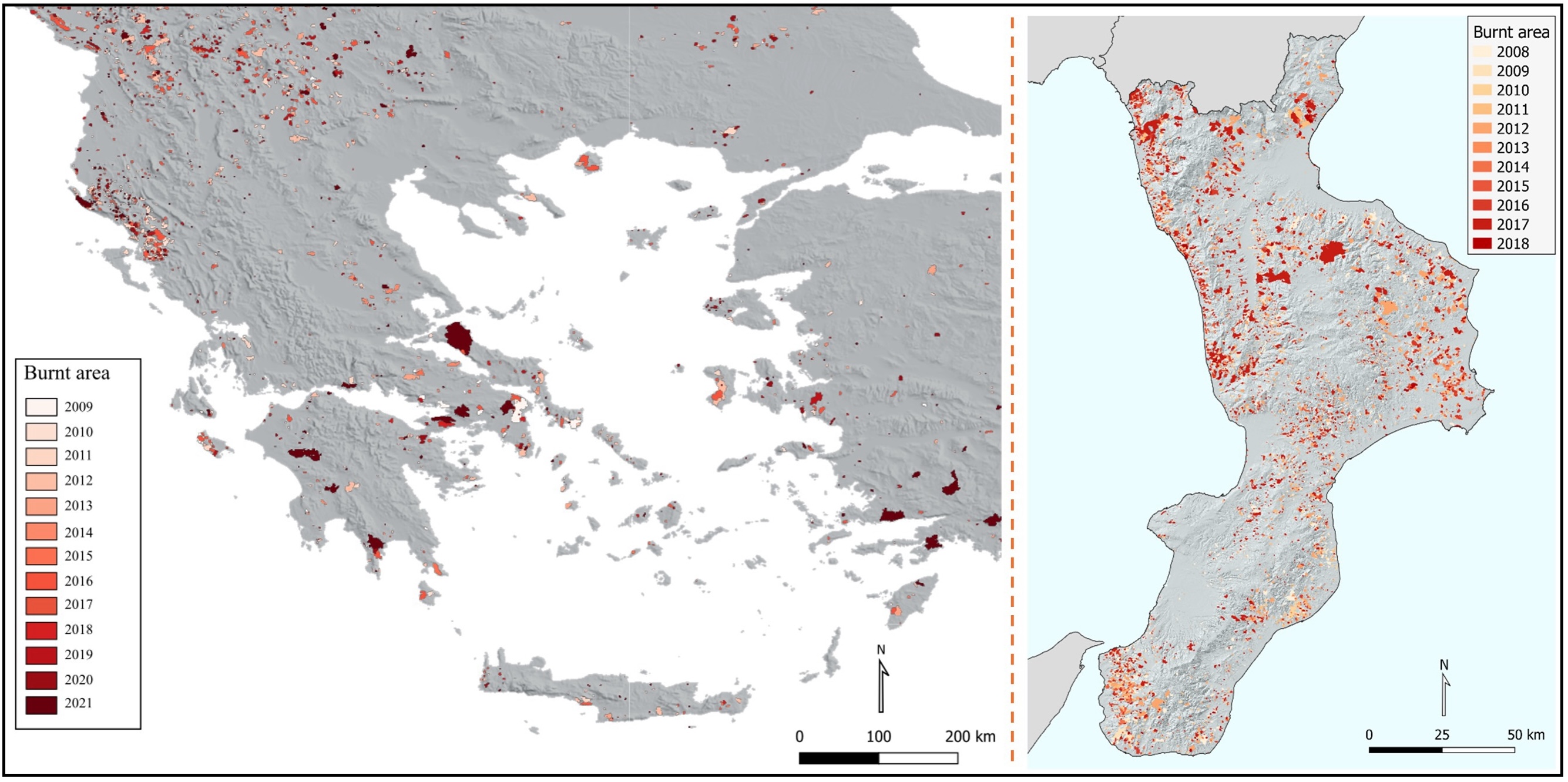}
\caption{Wildfire burnt areas, accoring to the datasets used in the experimentation and referring to the Greece and the southern Italy.}
\label{fig:burned_area}
\end{figure}

This paper moves a further step in the direction of defining more accurate wildfire indices, by proposing a predictive approach based on the \emph{Contrastive Learning} (CL) framework (see, \eg,~\citep{chen2020simple,khosla2020supervised,chuang2020debiased,wu2018unsupervised,henaff2020data,hjelm2018learning,chu2021cuco}). 
Additionally, we also aim to reduce the computational cost associated with the forecast forecasting task.
More specifically, our goal is to devise a supervised CL approach tailored to distinguish the severity of risks of the various geographical areas by suitably embedding such areas into the latent space: clusters of areas with similar kinds of wildfire risks should be drawn closer together, whereas areas exhibiting different behaviors should be driven apart. 
By employing a more principled method to design the internal structure of the latent space, we can acquire significant meaningful representations, even when the input information is reduced in size. Eventually, this will positively affect the total computational expense of the prediction by decreasing the number of necessary operations.
However, in practice, implementing this approach poses several challenges in such a given application scenario, where similar risks emerge from areas that can be very different from each others in terms of their features (see Figure~\ref{fig:burned_area}). Indeed, moving from the empirical observation that a basic supervised CL approach is ineffective to reach the prefixed goals, we adopt an ad-hoc architecture and introduce a carefully designed sampling strategy that leverages the dynamics of morphologically similar regions. 
Specifically, we examine the impact of incorporating a contrastive term during model training: first, we apply the contrastive term in a fine-tuning phase following the main training; second, we assess the results when the complete training includes this additional regularization term.
Subsequently, we suggest refining the label-based sampling approach by implementing a strategy that preserves morphological similarity among samples. We therefore propose two strategies, namely ``historical-based'' and ``curriculum-based'' samplings, to enhance the contrastive signal provided to the model during training.

The resulting methods has been implemented and its performances have been assessed on a number of real datasets. It emerged that the curriculm approach improves on the performances of current state-of-the-art methods, thereby providing 
a significant and practical contribution to the contrast of the phenomenon of wildfires. In fact, with respect to the latter perspective, it is worthwhile noticing that we also performed an ablation study specifically tailored to assess the impact on the quality of the predictions of the (geographical) area size considered for the embedding. The findings demonstrated that the proposed method shows a minor reduction in performance solely when the contextual information is significantly reduced, from $25\times25$ to $1\times1$ patch size. Consequently, by decreasing the input size, we can minimize the total computational cost of the prediction process, allowing multiple forecasts to be made using the same computational resources with the integration of updated dynamical information.

These advancements offer direct benefits to stakeholders involved in wildfire management, including public authorities, environmental agencies, and emergency response coordinators, by enabling more timely and resource-efficient interventions.

\section{Related works}\label{sec:related}

The application of deep learning methodologies to predict wildfire risk has attracted significant attention in recent years. This section provides a review of some relevant related works in this field and, in addition, it provides some background and references on \emph{Constrastive Learning}, for it being a key ingredient of our architecture.

\subsection{Predicting wildfire risk}

The influence of diverse factors on the incidence of forest fires has been firstly investigated by Wu \etal~\citep{Wu2021}. The work also compared the predictive performance of a multilayer perceptron (MLP) with that of logistic regression for wildfire prediction.
In the same year, the research detailed in~\citep{Pais2021} explored the impact of landscape topology—defined by the spatial distribution and interaction of various land-cover types—on fire ignition. 
This study introduced a deep learning model, Deep Fire Topology, which employs a convolutional neural network (CNN) to evaluate and predict the risk of wildfire ignition.

Hout \etal~\citep{Huot2022} subsequently conceptualized wildfire risk prediction as a scene classification challenge and employed U-Net architectures to anticipate wildfire propagation. 
A similar approach utilizing a U-Net++ model for global wildfire forecasting was presented by~\citep{prapas2022}. 

In the realm of recurrent neural networks, Yoon and Voulgaris~\citep{yoon2022} introduced a method leveraging a network with gated recurrent units (GRUs) to model historical data, complemented by a convolutional neural network (CNN) to predict wildfire probability maps over multiple temporal steps. 

Kondylatos \etal~\citep{Kondylatos2022} assessed various deep learning models, including long short-term memory (LSTM) and convolutional LSTM networks, demonstrating superior performance over traditional Fire Weather Index (FWI) methods in predicting next-day fire danger. 
Additionally, the study employed explainable AI techniques to analyze the critical influence of wetness-related variables.

Recently, \citep{eddin2023location} introduced a dual 2D-3D convolutional neural network (CNN) framework for predicting wildfire risks. 
The 2D CNN component processes static features, including digital elevation, slopes, road proximity, population density, and proximity to water bodies. 
In contrast, the 3D CNN component handles dynamic factors such as temperature, diurnal land surface temperatures, soil moisture levels, relative humidity, wind velocity, 2-meter air temperature, NDVI, atmospheric pressure, 2-meter dewpoint temperature, and precipitation totals.
Furthermore, two adaptive normalization blocks, sensitive to local positioning, were integrated into the 3D CNN branch to adjust dynamic features using static features. 
Empirical evaluations on the FireCube \citep{prapas2022firecube} and NDWS (Next Day Wildfire Spread) \citep{huot2022next} datasets revealed that this method surpassed traditional approaches such as random forest, XGBoost, LSTM, and convLSTM in performance. 
To the best of our knowledge, it also establishes a new state of the art on these datasets. Consequently, we adopt this model as the foundational architecture and benchmark for our study.

\subsection{Contrastive learning}

Contrastive learning is commonly referred to loss functions originating from metric distance learning or triplet-based approaches~\citep{chopra2005learning,weinberger2009distance,schroff2015facenet}. 
These functions are employed to enhance representation learning, usually within supervised settings where labels guide the selection of positive and negative pairs. 
The primary distinction between triplet losses and contrastive losses pertains to the number of positive and negative pairs associated with each data point; specifically, triplet losses engage exactly one positive and one negative pair for each anchor.
When positive pairs are derived from the same class, selecting negative samples becomes more complex. Schroff \etal \citep{schroff2015facenet} emphasized the necessity of meticulous negative mining to attain optimal performance.

Self-supervised contrastive losses similarly employ a single positive pair for each anchor sample. 
These pairs are identified either through co-occurrence \citep{henaff2020data,hjelm2018learning,tian2020contrastive} or data augmentation \citep{chen2020simple}, while numerous negative pairs are associated with each anchor. 
Typically, these negatives are chosen randomly and uniformly, leveraging weak knowledge such as patches from disparate images or frames from randomly chosen videos. This method presumes a minimal likelihood of false negatives.

In recent developments, Khosla \etal \citep{khosla2020supervised} suggested incorporating multiple positives per anchor in addition to numerous negatives. 
This approach of using numerous positive and negative pairs for each anchor has enabled the authors to achieve state-of-the-art performance without the need for challenging hard negative mining, which can be difficult to fine-tune. 
In this regard, sampling good candidates plays a crucial role during training as documented in~\citep{xu2022negative}. 
Robinson \etal~\citep{robinson2020contrastive} study how to sample good and informative negative examples for CL. They propose a new unsupervised method to select the hard-negative samples with user control. 
Jiang \etal~\citep{jiang2021improving} propose a principled approach to strategically select unlabeled data from an external source, in order to learn generalizable, balanced and diverse representations for relevant classes. 
Lastly, the authors in~\citep{chu2021cuco} introduce a curriculum CL framework that incrementally selects negative samples ranging from easy to challenging, using a score function to identify the hardness of the negatives.

In our study, we maintain the use of multiple positive and negative samples for each anchor, and design two different sampling strategies on the labeled data. However, we differentiate our method by restricting the selection
along the temporal dimension and the morphological similarity. 
As discussed in the following section, the historical-based sampling solely uses temporally varied versions of the same patch as positive and negative samples. In contrast, the curriculum-based approach involves sampling patches with static features, such as morphology and trigger factors, that increasingly differ from those of the anchor patch.

\section{Datasets description}\label{sec:dataset}

We selected two datasets for our study. 
The first is the FireCube dataset~\citep{prapas2022firecube} which is intended for pixel-wise classification and documents wildfire events in Greece over a decade. To offer a detailed and current representation of wildfire occurrences, it integrates diverse data sources, including satellite imagery, meteorological data, and historical fire records. 
Daily updates ensure the dataset remains current, enhancing the accuracy and reliability of wildfire predictions and response strategies. FireCube thus constitutes a substantial advancement in wildfire research and management.

This dataset encompasses an area of $1253\text{ km} \times 983\text{ km}$ in the Eastern Mediterranean. The key aim is to forecast wildfire occurrences exceeding $0.3~{\text{km}}^2$ in each cell for the subsequent day, within a binary classification framework where the positive class indicates fire occurrence. 
We utilized variables suggested by Kondylatos et al.~\citep{Prapas2}. As discussed in Section~\ref{sec:method}, to address dataset imbalance, we applied a sampling method to achieve balance~\citep{Prapas2}. 
Post-sampling, the dataset comprises $71471$ training examples ($13518$ positive, $57953$ negative from $2009$ to $2018$), $6430$ validation cases ($1300$ positive, $5130$ negative for $2019$), and $42820$ test cases ($1228$ positive, $4860$ negative for $2020$, and $4407$ positive, $32325$ negative for $2021$). Notably, 2021's test data capture a significant fire season in Greece~\citep{Prapas2}. The data samples can be accessed in~\citep{prapas_ioannis_2022_6528394}.

\begin{figure}[ht]
  \centering
  \includegraphics[width=8.5cm]{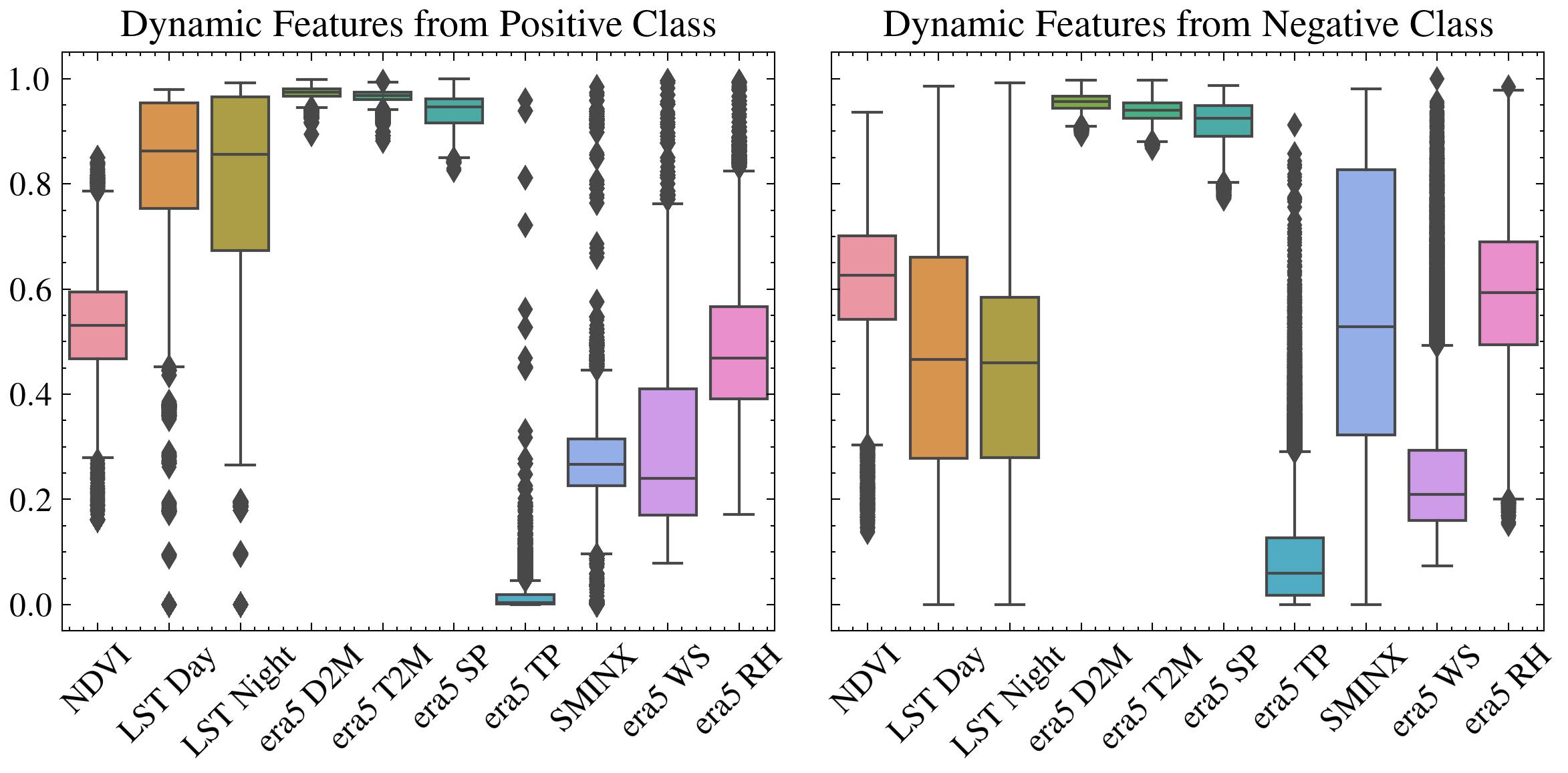}
  \caption{Box plot of the dynamic features from the balanced version of the Greece dataset.}
  \label{fig:box_dyn_greece}
\end{figure}

Figure~\ref{fig:box_dyn_greece} illustrates the distribution of data values across the two classes, focusing solely on the dynamic features of the balanced Greece dataset. 

The second dataset focuses on a specific region in southern Italy, namely Calabria, a peninsula of approximately $15,000 \text{ km}^2$ with a predominantly north-south orientation and a coastline of about $800 \text{ km}$. 
The fire events were directly detected through field observations by the Carabinieri Forestale, the official police authority responsible for this task. Consequently, this dataset offers a significantly higher level of accuracy and detail.
The dataset is derived from spatial interpolation of meteorological variables from station measurements provided by the Regional Monitoring Network. 
This dataset also has a higher spatial resolution, with a cell size of 100 meters, in contrast to the $1 \text{ km}$ resolution of the FireCube dataset. 
To reduce the number of patches due to the higher resolution, a fixed-size grid partitioning approach is employed to produce non-overlapping patches. 
However, this method does not ensure that wildfire occurrences are centered within the patches, requiring the model to learn a more complex task in this new configuration.
Thus, the Calabria dataset comprises a total of $23079$ training samples ($8666$ positive and $15453$ negative from years $2008-2015$), $3049$ validation samples ($1159$ positive and $1887$ negative for the year 2016), and $6380$ testing samples ($2239$ positive and $4141$ negative for the years $2017-2018$).

\begin{figure}[!ht]
  \centering
  \includegraphics[width=8.5cm]{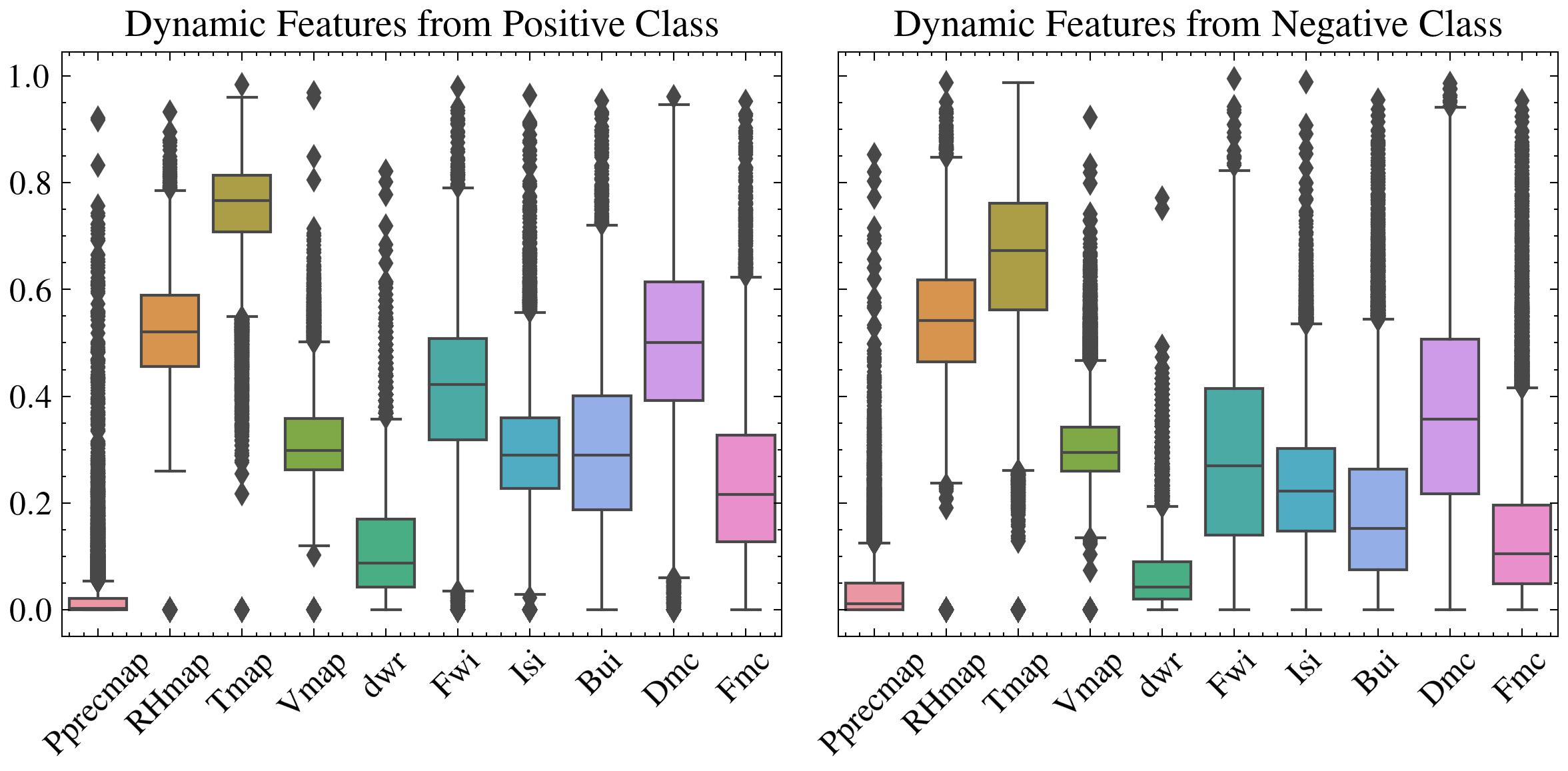}
  \caption{Box plot of the dynamic features from the Calabria dataset.}
  \label{fig:box_dyn_cal}
\end{figure}

Figure~\ref{fig:box_dyn_cal} illustrates the distribution of data values across the two classes of the balanced Calabria dataset. 

\section{Method}\label{sec:method}

In this section, we describe our methodology for enhancing the accuracy of wildfire risk predictions. This improvement is accomplished by refining the latent representations produced by the model through a tailored CL strategy. 
Due to the high dimensionality and complexity of the data, a direct application of CL would be ineffective. 
Consequently, to manage the intricacies of multivariate spatio-temporal data, we introduce a two-stage approach. 
The first phase involves selecting a representative subset from the original dataset for training a deep neural network. In the second phase, the network is trained using supervised learning. 
The sampling strategy may either leverage the contrastive signal (via label or curriculum sampling) throughout training or divide training into two sub-phases: initial training on the full dataset without sampling, followed by fine-tuning with historical sampling.
In either case, the CL methods aim to enhance the latent representations of the dynamic features to improve the accuracy and reliability of the prediction.
We now formalize the data context and proceed to elaborate on each of the stages in detail. 

The conceptual framework represents the application scenario as a multivariate spatio-temporal data cube, structured with two spatial dimensions for geographical coverage and a third for sequential temporal observations. 
Each entry contains multiple feature values, with the overall region divided into smaller subdivisions, or \textit{patches}, to define data granularity.
A patch is defined as a discrete geographic area, spanning from a few hundred square meters to several kilometers. 
These patches are typically described by static and dynamic attributes. 
Formally, we can represent the data associated with a patch as a tuple $ x = \langle v, t, w, h \rangle $, where $ v $ encapsulates the features (both static and dynamic), $ t $ represents the temporal dimension ranging from $ t = 0 $ to $ T $, and $ w $ and $ h $ correspond to the spatial dimensions.

After formalizing the patches, we propose two labeling strategies for binary classification. 
Using a sliding-window approach, the area is segmented into overlapping patches, where the labeling is determined by the central cell of each patch. Here, a positive label indicates that a wildfire will ignite at the center of the $w \times h$ area at time $t+1$. 
Alternatively, using a grid-based, non-overlapping patch system, a positive label denotes any wildfire occurrence within the patch at time $t+1$. While this method increases prediction complexity, it addresses high-resolution data issues and helps reduce computational costs. 

Regardless of the labeling strategy, the dataset remains imbalanced, with fewer positive samples. Our initial methodology phase is designed to address this imbalance.

\subsection{Phase One}

In the initial stage, the primary step involves structuring the raw data into patches. 
The process of creating these patches is defined with the patch size parameter, which regulates the amount of contextual information each patch contains. 
Employing larger patch sizes allows the model to capture more information, albeit at the risk of introducing more varied data, which might negatively impact the model's performance while uselessly increasing the computational cost of the forecast. 

Following the creation and labeling of patches, it becomes necessary to select a subset of negative samples due to the inherent data imbalance. 
The number of negative samples should ideally match that of positive ones, though handling a slightly higher number remains feasible. 
Notably, significant discrepancies in feature values between samples of the two classes can simplify the classification task, as the model could learn to discriminate based on non-informative features unrelated to wildfire events. 
To mitigate this, a similarity-based subsampling method is advisable.

Initially, an informative static feature, such as the Land Use Susceptibility Index, is selected as a proxy to evaluate the similarity between patches. 
Subsequently, an appropriate number of bins is determined for partitioning the negative samples. For each positive sample in the dataset, one or more negative samples are drawn from the bin corresponding to the specific feature value of the positive sample. 
This approach ensures the creation of a pseudo-balanced version of the original dataset.
We notice that while a basic static feature is currently used as a proxy for measuring similarity, future methodologies could incorporate more advanced techniques, such as clustering methods based on embeddings generated by specialized neural networks.

The final step in the initial phase addresses the temporal aspect of the data. The temporal parameter, denoted as $ t $, defines the extent of historical information accessible to the model for making predictions. 
The optimal value of $ t $ can be determined through hyper-parameter tuning, especially in response to the variability of certain features. Empirical evidence presented in~\citep{eddin2023location} indicates that incorporating data from the preceding ten time steps yields satisfactory results.

\subsection{Phase Two}

In the second phase, a neural network of adequate size is trained on the updated version of the dataset generated in the first phase. 
Conceptually, any chosen model should be capable of effectively managing the two types of features. 
For instance, Eddin et al.~\citep{eddin2023location} propose a dual-branch architecture that processes dynamic and static features independently while allowing static features to influence dynamic features through a normalization signal.
In general, we can define any parametric function $f_\theta$ that accepts as input the outputs of two distinct, yet potentially interrelated functions $f_{\theta}^{d}$ and $f_{\theta}^{s}$. 
These functions map dynamic and static features to their respective latent spaces, resulting in latent vectors $z_d$ and $z_s$. 
Depending on the complexity of the dataset, these representations might require varying degrees of additional non-linear transformations before computing the classification output.

The predicted value $\hat{y}=f_{\theta}(z_d,z_s)$, where $z_d=f_{\theta}^{d}(x_d)$ and $z_s=f_{\theta}^{s}(x_s)$, is calculated for the patch $x=\langle x_d, x_s \rangle$%
\footnote{Here, $x_d$ and $x_s$ refer to the sets of dynamic and static variables associated with the patch $x$}. 
This predicted value is subsequently used to train the entire model in a supervised manner, utilizing an appropriate objective function such as cross-entropy loss $\mathcal{L}_{CE}(y, \hat{y})$, where $y$ is the label assigned to the patch $x$.

The purpose of CL is therefore to improve the informative value of dynamic feature representations, denoted as $z_d$. 
Depending on the chosen sampling strategy we can either use the contrastive signal during the entire training (labels or curriculum sampling), or  splitting the training into two parts and training on the full dataset without CL, then fine-tuning with historical sampling and CL.
This differentiation is essential due to the significant reduction in data accessible for training caused by historical sampling, as we will discuss later. 

The emphasis on $z_d$ stems from the understanding that dynamic features are the primary source of information for the classification task. 
Consequently, improving the quality of their latent representations is likely to result in more accurate and reliable predictions.
Additionally, since our approach operates within a supervised framework, we can leverage the label information of an anchor sample $x^a$ to identify positive and negative samples, $x^p$ and $x^n$ respectively. 
This allows us to apply a suitable loss function, such as the triplet margin loss~\citep{chechik2010large}, as an auxiliary regularization term. This loss function is defined as follows:

\small 
\begin{equation}\label{eq:tri_loss}
    \begin{split}
    \mathcal{L}_{TL}(z_d^a, z_d^p, z_d^n) & = \\ 
    & \hspace{-15mm}\mathbb{E}_{z_d^p, z_d^n} \left[ \max\{d(z_d^a, z_d^p) - d(z_d^a, z_d^n) + m, 0\} \right]
    \end{split}
\end{equation}

In Eq. (\ref{eq:tri_loss}), the latent representation $z_d^a = f_{\theta}^{d}(x^a)$ corresponds to the anchor sample $x^a$, and $z_d^p = f_{\theta}^{d}(x^p)$ is derived by sampling a positive instance $x^p \sim P$ using $x^a$'s label information from the entire dataset. 
Here, $P$ denotes the set of samples with the same label as $x^a$. A similar process is applied to obtain $z_d^n$. The function $d(z_d^i, z_d^j) = ||z_d^i - z_d^j||_p$ is the chosen $p$-norm, and $m$ denotes a predetermined margin.

In our experiments, we also investigate a second constrastive term introduced in~\citep{chen2020simple,khosla2020supervised}, named \textit{Supervised Contrastive Loss}. 
Pursuing the same goal of the previous loss, it computes the pairwise similarities between all the latent projections in a batch, scaled by a temperature parameter to control the sharpness of the distribution. 
For each sample, it identifies positive pairs (the other samples in the batch that share the same class label) and calculates the negative log-probability of these positive pairs relative to all other samples, excluding the anchor itself, to avoid trivial solutions. It can be defined as follows:

\small \begin{equation}\label{eq:scl_loss}
  \mathcal{L}_{SCL}
  =\frac{1}{B} \sum_{i=1}^{B} \frac{1}{|P(i)|} \sum_{j \in P(i)} -\log \frac{\exp( \frac{\mathbf{z}_d^{i\top} \cdot \mathbf{z}_d^j}{\tau})}{\sum\limits_{k=1,\, k \ne i}^{B} \exp( \frac{\mathbf{z}_d^{i\top} \cdot \mathbf{z}_d^k}{\tau}) }
\end{equation}

where $B$ is the batch size, $P(i)$ the set of indices of all positive samples for the anchor sample $i$ according to its label and $\tau$ the temperature parameter that controls the scaling of the similarities computed by the dot product. 

In our proposed sampling strategy, we opted for using a triplet-loss, as it promises improved optimization of relative distances and higher discriminative capability by employing a margin.
In general, contrastive loss functions on pairwise comparisons aiming to reduce the distance between an anchor and a positive example while increasing the distance between the anchor and a negative example. However, it does not rigorously ensure that the negative example is adequately distant from the anchor compared to the positive example.
In contrast, triplet-loss engages with a triplet of samples and should guarantee that the distance between the anchor and the positive example is less than the distance between the anchor and the negative example by at least by a specified margin.

The objective function used for the training is then defined as:
\begin{equation}
    \mathcal{L}_{CE}(y,\hat{y})+\gamma * \mathcal{L}_{CL}(z_d^a,z_d^p,z_d^n)
\end{equation}

where \small{$\mathcal{L}_{CE}$} represents the binary cross-entropy, \small{$\mathcal{L}_{CL}$} is one between \small{$\mathcal{L}_{TL}$} and \small{$\mathcal{L}_{SCL}$}, and \small{$\gamma=|\mathcal{L}_{CE}| /|\mathcal{L}_{CL}|$} for \small{$|\mathcal{L}_{CL}|>0$}, and $0$ otherwise. This $\gamma$ adequately scales the contribution of the contrastive term to match the magnitude of the primary target of the learning which is \small{$\mathcal{L}_{CE}$}.

\begin{table*}[ht]
    \caption{{Average Difference over Dynamic Features from the FireCube dataset computed using triplets chosen solely based on label data, triplets selected through our historical methodology, and finally through our curriculum methodology. 
    }}
    \centering
    \resizebox{.85\linewidth}{!}{
    \begin{tabular}{crccc|ccc|ccc}
    & \multicolumn{4}{c}{\textit{Avg. Anchor-Positive Diff $(\downarrow)$}} & \multicolumn{3}{c}{Avg. Anchor-Negative Diff $(\uparrow)$}  & \multicolumn{3}{c}{\textit{ratio $(\uparrow)$}}\\
     Feature Resolution & Feature Name & \textit{Label} & \textit{Historical} & \textit{Curriculum}  & \textit{Label} & \textit{Historical }  & \textit{Curriculum} & \textit{Label} & \textit{Hist.}  & \textit{Curr.} \\
    \midrule
\multicolumn{1}{c}{\multirow{3}{*}{\textit{High}-$1$Km}}  
& NDVI \tiny{1 km 16 days} & 0.45 \tiny{ $\pm$ 1e-01} & 0.27 \tiny{ $\pm$ 8e-02} & 0.16 \tiny{ $\pm$ 8e-02} & 0.43 \tiny{ $\pm$ 1e-01} &0.32 \tiny{ $\pm$ 8e-02} &0.29 \tiny{ $\pm$ 9e-02} &1.0 & 1.2 & \textbf{1.8} \\
& LST Day 1km & 0.42 \tiny{ $\pm$ 1e-01} & 0.32 \tiny{ $\pm$ 2e-01} & 0.27 \tiny{ $\pm$ 2e-01} & 0.54 \tiny{ $\pm$ 1e-01} &0.55 \tiny{ $\pm$ 1e-01} &0.40 \tiny{ $\pm$ 1e-01} &1.3 & \textbf{1.7} & 1.5 \\
 & LST Night 1km & 0.43 \tiny{ $\pm$ 1e-01} & 0.35 \tiny{ $\pm$ 2e-01} & 0.21 \tiny{ $\pm$ 1e-01} & 0.54 \tiny{ $\pm$ 1e-01} &0.53 \tiny{ $\pm$ 2e-01} &0.39 \tiny{ $\pm$ 1e-01} &1.3 & 1.5 & \textbf{1.8} \\
 \midrule
\multicolumn{1}{c}{\multirow{7}{*}{\textit{Low}-$9$Km}} 
& era5 max d2m & 0.10 \tiny{ $\pm$ 5e-02} & 0.11 \tiny{ $\pm$ 9e-02} & 0.03 \tiny{ $\pm$ 2e-02} & 0.16 \tiny{ $\pm$ 6e-02} &0.26 \tiny{ $\pm$ 1e-01} &0.04 \tiny{ $\pm$ 2e-02} &1.6 & \textbf{2.3} & 1.4 \\
 & era5 max t2m & 0.09 \tiny{ $\pm$ 5e-02} & 0.11 \tiny{ $\pm$ 9e-02} & 0.04 \tiny{ $\pm$ 3e-02} & 0.20 \tiny{ $\pm$ 7e-02} &0.47 \tiny{ $\pm$ 1e-01} &0.07 \tiny{ $\pm$ 3e-02} &2.2 & \textbf{4.2} & 1.8 \\
 & era5 max SP & 0.21 \tiny{ $\pm$ 8e-02} & 0.20 \tiny{ $\pm$ 1e-01} & 0.02 \tiny{ $\pm$ 1e-02} & 0.23 \tiny{ $\pm$ 7e-02} &0.21 \tiny{ $\pm$ 1e-01} &0.03 \tiny{ $\pm$ 1e-02} &1.1 & 1.1 & \textbf{1.8} \\
 & era5 max TP & 0.07 \tiny{ $\pm$ 6e-02} & 0.03 \tiny{ $\pm$ 5e-02} & 0.06 \tiny{ $\pm$ 6e-02} & 0.14 \tiny{ $\pm$ 9e-02} &0.27 \tiny{ $\pm$ 2e-01} &0.08 \tiny{ $\pm$ 7e-02} &2.0 & \textbf{9.3} & 1.2 \\

 & era5 max Wind Speed & 0.21 \tiny{ $\pm$ 1e-01} & 0.16 \tiny{ $\pm$ 1e-01} & 0.09 \tiny{ $\pm$ 9e-02} & 0.21 \tiny{ $\pm$ 1e-01} &0.16 \tiny{ $\pm$ 1e-01} &0.15 \tiny{ $\pm$ 1e-01} &1.0 & 1.0 & \textbf{1.7} \\
 & era5 min RH & 0.31 \tiny{ $\pm$ 1e-01} & 0.21 \tiny{ $\pm$ 1e-01} & 0.19 \tiny{ $\pm$ 1e-01} & 0.36 \tiny{ $\pm$ 1e-01} &0.41 \tiny{ $\pm$ 2e-01} &0.32 \tiny{ $\pm$ 2e-01} &1.2 & \textbf{2.0} & 1.6 \\
 & SMINX & 0.31 \tiny{ $\pm$ 1e-01} & 0.15 \tiny{ $\pm$ 1e-01} & 0.27 \tiny{ $\pm$ 2e-01} & 0.47 \tiny{ $\pm$ 2e-01} &0.49 \tiny{ $\pm$ 1e-01} &0.39 \tiny{ $\pm$ 2e-01} &1.5 & \textbf{3.2} & 1.5 \\
\bottomrule
    \end{tabular}
    }
    \label{tab:delta_features_greece}
\end{table*}

\subsection{Sampling strategies}

The main complexity in implementing the CL approach in this context stems from the significant variation in dynamic features among patches sharing the same labels, attributable to inherent differences in the nature of the areas covered (refer to Table~\ref{tab:delta_features_greece}). 
Consequently, the model must reconcile these disparities within closely related latent representations. 
Experimental results indicate that this leads to the acquisition of noisier latent representations, thereby reducing the overall model performance.

To mitigate this, we propose restricting the sampling following two distinct approaches: historical and curriculum sampling. 
The historical sampling limits the sampling to patches within the history of the anchor or its closest neighbors\footnote{We had to include closest neighbors due to the scarcity of different versions of positive patches in the studied datasets.}. 
It is worth to notice that, with this strategy, we are focusing solely on patches with positive events to construct historical sets, we thus significantly reduce the volume of data available for training. 
Without appropriate countermeasures, this reduction could negatively impact the training process and result in suboptimal performance.

Thus, we also propose a curriculum-based strategy to sample patches according to their morphological similarity to the anchor. 
We use the term \textit{curriculum} for this sampling strategy, since we progressively sample patches that are different from the anchor using a score function $f_{score}(x_s^i, x_s^j)$.
We implement this function as the L2-norm between the normalized version of $x_s^i$ and $x_s^j$.
The primary advantage of employing similarity-based sampling lies in its ability to leverage the entire dataset for the construction of positive and negative sample pairs.
Additional details in Appendix A of the Supplementary Material~\citep{loscudo2025}.

Either of the above approaches limit the variability among input features, allowing the model to learn smoother $z_d$ representations, as shown in Table~\ref{tab:delta_features_greece} for the FireCube dataset and Table 5 in the Supplementary Material for the Calabria dataset~\citep{loscudo2025}. 
Those tables report the average normalized difference for dynamic features, calculated using triplet-based comparisons.
%
For each anchor patch, we thus randomly select ten positive and ten negative samples using the label-based sampling, while for the historical and curriculum sampling we create ten triplets, respectively, using pre-computed maps. 

The results show that, for higher-resolution features, the tighter constraints of historical and curriculum sampling produce the larger ratio between the mean distance between the dynamic features of the anchor and the negative samples and the anchor and the positive samples, $\delta(x_d^a,x_d^n)/\delta(x_d^a,x_d^p)$.
Those features should provide more useful information than the lower ones for which the historical still maintains a high ratio in general, whereas the curriculum pays a small price due to the higher numerosity of the samples.
The random label-based sampling reaches a good difference between the anchor and the negative samples, but shows a similar variability also between the anchor and the positive ones.
Finally, our curriculum sampling shows the lowest difference on the anchor-positive pairs, but also reports the smallest difference among the comparisons between anchors and negative samples.

We evaluate the proposed contrastive sampling strategies within two distinct training paradigms.
In the first approach, CL is applied as a fine-tuning step after the model has been pre-trained on the full dataset. At this stage, the model has already learned discriminative features in the $z_d$ representations, primarily due to the diversity of negative samples in the balanced training set. The CL objective is then used to refine these representations using a more selectively curated dataset.
In the second approach, the model is trained end-to-end with the contrastive objective from the outset. Further details are provided in Appendix B of the Supplementary Material~\citep{loscudo2025}.

\section{Results}
\label{sec:results}

We conduct experiments on the training methodologies delineated in Section~\ref{sec:method} with dual objectives. 
The primary objective is to evaluate the effectiveness of our sampling techniques in terms of classification accuracy, comparing it against various models and different CL sampling techniques. 
The secondary aim is to examine the influence of the geographical area size on the quality of the predictions. 

In the CL framework, we evaluate four distinct configurations: the triplet-marginal loss approach Eq.~(\ref{eq:tri_loss}) using standard label-based sampling, alongside our historical and curriculum sampling methods, and the modern Supervised Contrastive Loss Eq.~(\ref{eq:scl_loss}) using label-based sampling.

We then examine two potential classification frameworks. 
In the first framework, we model the dynamics of the central cell within a specified area using all adjacent cells as sources of contextual information, the FireCube dataset. 
Conversely, in the second framework, we aim to model the dynamics of all points within the area, thus requiring the model to accommodate a more complex data distribution, the Calabria dataset.

To evaluate the influence of contextual information on prediction accuracy, we conducted experiments using various patch sizes. 
Beginning with the $25\times25$ patch size as utilized in~\citep{eddin2023location}, we progressively decreased the dimensions to define three additional scenarios, maintaining fixed the center cell: $15\times15$, $5\times5$, and $1\times1$. 
For each specified patch size, all models were retrained from the initial state.

In this study, the model \textit{LOAN} introduced in~\citep{eddin2023location} serves as the reference baseline, modified slightly to fit smaller patch sizes.
These architectural modifications are consistently employed across all models implementing CL methodologies.

We also select two recent larger models that employ the self-attention mechanism to capture spatiotemporal dependencies. 
We perform a comparative analysis with recent transformer-based models, namely TimeSformer~\citep{bertasius2021space} and Video Swin Transformer 3D~\citep{liu2022video}. 
These transformer models offer a distinct advantage over CNN-based model by reducing the reliance on strong inductive biases, allowing them to generalize better to diverse spatio-temporal dynamic patterns. 
However, this flexibility comes at a cost: transformers typically demand significantly higher computational resources for training compared to CNNs.
The CNN model utilized, for instance, has approximately $414k$ parameters, while TimeSformer has around $1.16M$, and the Swin Transformer is the most extensive with $1.8M$ parameters.
This trade-off between flexibility and computational efficiency is an important factor when considering transformer models for forecasting tasks, especially in resource-constrained environments.


\paragraph{Experimental settings details.}
\label{app:experimental_settings}

In accordance with the methodology presented by Eddin et al.~\cite{eddin2023location}, our model training encompassed a total of 40 epochs, maintaining all architectural parameters and hyperparameters consistent with the original study. 
The sole modification in our contrastive learning (CL) approaches involved an increase in the learning rate from the initial $3 \times 10^{-5}$ to $3 \times 10^{-4}$ (according to our experimental findings, our attempts to increase the learning rate in the original configuration resulted in a diminished performance).

As detailed in Section~\ref{sec:method}, this serves as a fine-tuning phase; thus, training with the CL term is initiated only after completing 30 epochs, followed by an additional $10$ epochs.
Using the triplet loss, for each batch item, regarded as an anchor, pairs of positive and negative samples are selected based on the information on the label. After experimental testing, we fix the margin value at $5$ for our
strategies and $20$ for the traditional label-based approach; the related ablation study is reported in Table 9 in the Supplementary Material~\citep{loscudo2025}.

To address the presence of negative samples in the label-based contrastive learning (CL) process, we limit the number of fine-tuning epochs to five. This approach ensures that each positive sample is encountered twice, similar to the historical contrastive sampling (CS) method: once as an anchor sample and once as a negative sample.

In the conventional CL framework based on the triplet-loss, the entire dataset is leveraged.  

Our historical CL method follows an analogous training regime to the label-based CL strategy but uses a subset of the initial dataset. 
Initially, we select all positive examples from the dataset. For each patch, two sets of positive and negative samples, derived from the patch's history, are constructed. During training, for each patch in the batch, positive and negative samples are randomly drawn from its historical data. 

Finally, we evaluate the newer supervised contrastive loss, as introduced in~\cite{khosla2020supervised}, as a substitute for the triplet loss. Unlike before, sampling is unnecessary; instead, every sample in the batch is used to calculate the contrastive loss. As for the historical case, the fine-tuning lasts for $10$ epochs.

In addition to our proposed approach, we perform a comparative analysis with recent transformer-based models, namely TimeSformer~\cite{bertasius2021space} and Video Swin Transformer 3D~\cite{liu2022video}. 
Both models leverage the self-attention mechanism to capture spatio-temporal dependencies within video data. The TimeSformer model utilizes divided space-time attention, enabling efficient modeling of long-range dependencies, while the Video Swin Transformer employs hierarchical attention mechanisms that improve feature extraction at multiple scales. 
These transformer models offer a distinct advantage over CNN-based model by reducing the reliance on strong inductive biases, allowing them to generalize better to diverse spatio-temporal dynamic patterns. 
However, this flexibility comes at a cost: transformers typically demand significantly higher computational resources for training compared to CNNs.
The CNN model utilized, for instance, has approximately $414k$ parameters, while TimeSformer has around $1.16M$, and the Swin Transformer is the most extensive with $1.8M$ parameters.
This trade-off between flexibility and computational efficiency is an important factor when considering transformer models for forecasting tasks, especially in resource-constrained environments.

All experiments were carried out using a single node with 96 CPUs, 512 GB RAM, and an NVIDIA V100 GPU with 16 GB VRAM.
Access to the code to replicate experiments 
can be granted upon request for academic purposes.

\paragraph{Greece Dataset.}

\begin{table}[ht]
  \caption{Aggregated results over the years 2020 and 2021 from the FireCube dataset are reported. Each value represents the mean performance across five independent trials. The best results are highlighted in \textbf{bold}. Class-wise performance details are provided in Appendix D of the Supplementary Material~\citep{loscudo2025}.}
  \centering
  \resizebox{.95\linewidth}{!}{
  \begin{tabular}{clc|c|c|c}
  \toprule
    PS & Model & \textit{Precision} & \textit{AUROC} & \textit{IoU} & \textit{F1}\\
    \midrule
\multirow{10}{*}{\rotatebox[origin=c]{90}{$1\times1$}} & 
TimeSformer & 0.90 \tiny{ $\pm$ 2e-02} & 0.95 \tiny{ $\pm$2e-03} & 0.82 \tiny{ $\pm$5e-03} & 0.90  \tiny{ $\pm$3e-03} \\
 & SwinTransformer3D & 0.89 \tiny{ $\pm$ 2e-02} & 0.95 \tiny{ $\pm$4e-03} & 0.80 \tiny{ $\pm$8e-03} & 0.89  \tiny{ $\pm$5e-03} \\
 & LOAN \small{(Baseline)} & 0.90 \tiny{ $\pm$ 5e-02} & 0.95 \tiny{ $\pm$1e-03} & 0.81 \tiny{ $\pm$1e-02} & 0.89  \tiny{ $\pm$7e-03} \\
  \cmidrule(r){2-6}
 & LOAN+\small{LTL - ft} & 0.91 \tiny{ $\pm$ 4e-02} & 0.97 \tiny{ $\pm$2e-03} & 0.83 \tiny{ $\pm$1e-02} & 0.91  \tiny{ $\pm$8e-03} \\
 & LOAN+\small{SCL - ft} & 0.90 \tiny{ $\pm$ 5e-02} & 0.95 \tiny{ $\pm$2e-03} & 0.81 \tiny{ $\pm$2e-02} & 0.90  \tiny{ $\pm$1e-02} \\
  & LOAN+\small{HTL - ft (Ours)} & 0.90 \tiny{ $\pm$ 1e-02} & 0.96 \tiny{ $\pm$8e-04} & 0.82 \tiny{ $\pm$5e-03} & 0.90  \tiny{ $\pm$3e-03} \\
  & LOAN+\small{CTL - ft (Ours)} & 0.93 \tiny{ $\pm$ 1e-02} & 0.98 \tiny{ $\pm$2e-03} & 0.87 \tiny{ $\pm$5e-03} & 0.93  \tiny{ $\pm$3e-03} \\
   \cmidrule(r){2-6}
 & LOAN+\small{LTL - full} & 0.91 \tiny{ $\pm$ 2e-02} & 0.97 \tiny{ $\pm$2e-03} & 0.84 \tiny{ $\pm$6e-03} & 0.91  \tiny{ $\pm$4e-03} \\
 & LOAN+\small{SCL - full} & 0.91 \tiny{ $\pm$ 3e-02} & 0.96 \tiny{ $\pm$4e-03} & 0.83 \tiny{ $\pm$1e-02} & 0.91  \tiny{ $\pm$8e-03} \\
  & LOAN+\small{CTL - full (Ours)} & \textbf{0.93} \tiny{ $\pm$ 2e-02} & \textbf{0.98} \tiny{ $\pm$1e-03} & \textbf{0.88} \tiny{ $\pm$4e-03} & \textbf{0.93}  \tiny{ $\pm$2e-03} \\
\midrule
\multirow{10}{*}{\rotatebox[origin=c]{90}{$5\times5$}} & 
TimeSformer & 0.88 \tiny{ $\pm$ 6e-02} & 0.95 \tiny{ $\pm$2e-03} & 0.77 \tiny{ $\pm$6e-02} & 0.87  \tiny{ $\pm$4e-02} \\
 & SwinTransformer3D & 0.91 \tiny{ $\pm$ 3e-02} & 0.96 \tiny{ $\pm$7e-03} & 0.83 \tiny{ $\pm$7e-03} & 0.91  \tiny{ $\pm$4e-03} \\
 & LOAN \small{(Baseline)} & 0.89 \tiny{ $\pm$ 8e-02} & 0.97 \tiny{ $\pm$2e-03} & 0.78 \tiny{ $\pm$4e-02} & 0.87  \tiny{ $\pm$2e-02} \\
  \cmidrule(r){2-6}
 & LOAN+\small{LTL - ft} & 0.90 \tiny{ $\pm$ 6e-02} & 0.97 \tiny{ $\pm$6e-03} & 0.80 \tiny{ $\pm$5e-02} & 0.89  \tiny{ $\pm$3e-02} \\
 & LOAN+\small{SCL - ft} & 0.91 \tiny{ $\pm$ 6e-02} & 0.97 \tiny{ $\pm$3e-03} & 0.82 \tiny{ $\pm$5e-02} & 0.90  \tiny{ $\pm$3e-02} \\
  & LOAN+\small{HTL - ft (Ours)} & 0.91 \tiny{ $\pm$ 2e-02} & 0.97 \tiny{ $\pm$1e-03} & 0.84 \tiny{ $\pm$5e-03} & 0.91  \tiny{ $\pm$3e-03} \\
  & LOAN+\small{CTL - ft (Ours)} & 0.94 \tiny{ $\pm$ 2e-02} & 0.99 \tiny{ $\pm$1e-03} & 0.89 \tiny{ $\pm$3e-03} & 0.94  \tiny{ $\pm$2e-03} \\
   \cmidrule(r){2-6}
 & LOAN+\small{LTL - full} & 0.89 \tiny{ $\pm$ 7e-02} & 0.97 \tiny{ $\pm$3e-03} & 0.79 \tiny{ $\pm$6e-02} & 0.88  \tiny{ $\pm$4e-02} \\
 & LOAN+\small{SCL - full} & 0.91 \tiny{ $\pm$ 5e-02} & 0.97 \tiny{ $\pm$4e-03} & 0.83 \tiny{ $\pm$4e-02} & 0.91  \tiny{ $\pm$2e-02} \\
  & LOAN+\small{CTL - full (Ours)} & \textbf{0.95} \tiny{ $\pm$ 2e-02} & \textbf{0.99} \tiny{ $\pm$2e-03} & \textbf{0.90} \tiny{ $\pm$1e-02} & \textbf{0.95}  \tiny{ $\pm$5e-03} \\
\midrule
\multirow{10}{*}{\rotatebox[origin=c]{90}{$15\times15$}} & 
TimeSformer & 0.89 \tiny{ $\pm$ 5e-02} & 0.95 \tiny{ $\pm$3e-03} & 0.79 \tiny{ $\pm$4e-02} & 0.88  \tiny{ $\pm$2e-02} \\
 & SwinTransformer3D & 0.91 \tiny{ $\pm$ 3e-02} & 0.96 \tiny{ $\pm$7e-03} & 0.83 \tiny{ $\pm$7e-03} & 0.91  \tiny{ $\pm$4e-03} \\
 & LOAN \small{(Baseline)} & 0.89 \tiny{ $\pm$ 8e-02} & 0.97 \tiny{ $\pm$2e-03} & 0.78 \tiny{ $\pm$4e-02} & 0.87  \tiny{ $\pm$3e-02} \\
  \cmidrule(r){2-6}
 & LOAN+\small{LTL - ft} & 0.89 \tiny{ $\pm$ 7e-02} & 0.97 \tiny{ $\pm$6e-03} & 0.79 \tiny{ $\pm$6e-02} & 0.88  \tiny{ $\pm$4e-02} \\
 & LOAN+\small{SCL - ft} & 0.90 \tiny{ $\pm$ 7e-02} & 0.97 \tiny{ $\pm$3e-03} & 0.80 \tiny{ $\pm$7e-02} & 0.89  \tiny{ $\pm$4e-02} \\
  & LOAN+\small{HTL - ft (Ours)} & 0.91 \tiny{ $\pm$ 3e-02} & 0.97 \tiny{ $\pm$1e-03} & 0.84 \tiny{ $\pm$6e-03} & 0.91  \tiny{ $\pm$3e-03} \\
 & LOAN+\small{CTL - ft (Ours)} & 0.94 \tiny{ $\pm$ 2e-02} & 0.99 \tiny{ $\pm$1e-03} & 0.89 \tiny{ $\pm$4e-03} & 0.94  \tiny{ $\pm$2e-03} \\
  \cmidrule(r){2-6}
 & LOAN+\small{LTL - full} & 0.90 \tiny{ $\pm$ 6e-02} & 0.96 \tiny{ $\pm$7e-03} & 0.80 \tiny{ $\pm$5e-02} & 0.89  \tiny{ $\pm$3e-02} \\
 & LOAN+\small{SCL - full} & 0.91 \tiny{ $\pm$ 6e-02} & 0.97 \tiny{ $\pm$2e-03} & 0.82 \tiny{ $\pm$5e-02} & 0.90  \tiny{ $\pm$3e-02} \\
  & LOAN+\small{CTL - full (Ours)} & \textbf{0.95} \tiny{ $\pm$ 3e-02} & \textbf{0.99} \tiny{ $\pm$2e-03} & \textbf{0.90} \tiny{ $\pm$1e-02} & \textbf{0.95}  \tiny{ $\pm$7e-03} \\
\midrule
\multirow{10}{*}{\rotatebox[origin=c]{90}{$25\times25$}} &
TimeSformer & 0.88 \tiny{ $\pm$ 5e-02} & 0.95 \tiny{ $\pm$3e-03} & 0.78 \tiny{ $\pm$3e-02} & 0.88  \tiny{ $\pm$2e-02} \\
 & SwinTransformer3D & 0.91 \tiny{ $\pm$ 3e-02} & 0.96 \tiny{ $\pm$1e-03} & 0.83 \tiny{ $\pm$7e-03} & 0.90  \tiny{ $\pm$4e-03} \\
 & LOAN \small{(Baseline)} & 0.91 \tiny{ $\pm$ 5e-02} & 0.97 \tiny{ $\pm$3e-03} & 0.83 \tiny{ $\pm$3e-02} & 0.91  \tiny{ $\pm$2e-02} \\
  \cmidrule(r){2-6}
 & LOAN+\small{LTL - ft} & 0.92 \tiny{ $\pm$ 5e-02} & 0.97 \tiny{ $\pm$4e-03} & 0.84 \tiny{ $\pm$2e-02} & 0.91  \tiny{ $\pm$1e-02} \\
 & LOAN+\small{SCL - ft} & 0.92 \tiny{ $\pm$ 3e-02} & 0.97 \tiny{ $\pm$5e-03} & 0.84 \tiny{ $\pm$1e-02} & 0.91  \tiny{ $\pm$6e-03} \\
  & LOAN+\small{HTL - ft (Ours)} & 0.91 \tiny{ $\pm$ 3e-02} & 0.97 \tiny{ $\pm$2e-03} & 0.84 \tiny{ $\pm$9e-03} & 0.91  \tiny{ $\pm$6e-03} \\
 & LOAN+\small{CTL - ft (Ours)} & \textbf{0.95} \tiny{ $\pm$ 3e-02} & 0.99 \tiny{ $\pm$1e-03} & 0.89 \tiny{ $\pm$6e-03} & 0.94  \tiny{ $\pm$3e-03} \\
 \cmidrule(r){2-6}
 & LOAN+\small{LTL - full} & 0.91 \tiny{ $\pm$ 7e-02} & 0.97 \tiny{ $\pm$5e-03} & 0.81 \tiny{ $\pm$6e-02} & 0.90  \tiny{ $\pm$4e-02} \\
 & LOAN+\small{SCL - full} & 0.92 \tiny{ $\pm$ 4e-02} & 0.97 \tiny{ $\pm$3e-03} & 0.85 \tiny{ $\pm$2e-02} & 0.92  \tiny{ $\pm$1e-02} \\
 & LOAN+\small{CTL - full (Ours)} & \textbf{0.95} \tiny{ $\pm$ 3e-02} & \textbf{0.99} \tiny{ $\pm$1e-03} & \textbf{0.91} \tiny{ $\pm$2e-02} & \textbf{0.95}  \tiny{ $\pm$1e-02} \\
\bottomrule
  \end{tabular}
  }
\label{tab:greek_results_tot_paper}
\end{table}

\begin{table}[ht]
    \caption{Average distance intra- and inter-class for latent codes calculated by the different models. We report the overall best results in \small{\textbf{black}} and the best results among the models fully trained with the contrastive terms in \small{\color{blue}{\textbf{blue}}}.}
    \centering
    \resizebox{1.\linewidth}{!}{
    \begin{tabular}{clc|c|c|c|c||c|c|c}
        & & & \multicolumn{4}{c}{\textit{Fine-tuning}} & \multicolumn{3}{c}{\textit{Full}}\\
        PS & Distance & Baseline & LTL & SCL & HTL & CTL & LTL & SCL & CTL\\
        \midrule
            \multirow{3}{*}{\rotatebox[origin=c]{90}{\tiny{$1\times1$}}}
        & \textit{Intra-Cl. $(\downarrow)$} & 1.07 & 0.89 & 1.13 & 0.91 & 1.06 & \textbf{0.86} & 1.12 & 1.02\\
        & \textit{Inter-Cl. $(\uparrow)$} & \textbf{1.42} & 1.27 & 1.4 & 1.43 & 1.39 & 1.23 & \btextbf{1.4} & 1.33\\
        & \textit{ratio $(\uparrow)$} & 1.33 & 1.42 & 1.24 & \textbf{1.58} & 1.32 & \btextbf{1.43} & 1.24 & 1.31\\
        \midrule
        \multirow{3}{*}{\rotatebox[origin=c]{90}{\tiny{$5\times5$}}}
        & \textit{Intra-Cl. $(\downarrow)$} & 0.86 & 0.83 & 1.08 & \textbf{0.8} & 0.93 & 0.86 & 1.08 & 0.9\\
        & \textit{Inter-Cl. $(\uparrow)$} & 1.16 & 1.21 & \textbf{1.35} & 1.26 & 1.27 & 1.16 & \btextbf{1.33} & 1.27\\
         & \textit{ratio $(\uparrow)$} & 1.35 & 1.46 & 1.25 & \textbf{1.58} & 1.36 & 1.35 & 1.23 & \btextbf{1.4}\\
        \midrule
        \multirow{3}{*}{\rotatebox[origin=c]{90}{\tiny{$15\times15$}}}
        & \textit{Intra-Cl. $(\downarrow)$} & 0.74 & \textbf{0.7} & 1.07 & \textbf{0.7} & 0.82 & \btextbf{0.71} & 1.03 & 0.8\\
        & \textit{Inter-Cl. $(\uparrow)$} & 1.03 & 1.14 & \textbf{1.34} & 1.17 & 1.17 & 1.04 & \btextbf{1.26} & 1.18\\
         & \textit{ratio $(\uparrow)$} & 1.39 & 1.62 & 1.25 & \textbf{1.67} & 1.42 & 1.46 & 1.23 & \btextbf{1.48}\\
        \midrule
        \multirow{3}{*}{\rotatebox[origin=c]{90}{\tiny{$25\times25$}}}
        & \textit{Intra-Cl. $(\downarrow)$} & 0.48 & 0.68 & 0.88 & 0.65 & 0.62 & 0.72 & 1.02 & \textbf{0.56}\\
        & \textit{Inter-Cl. $(\uparrow)$} & 0.75 & 1.17 & \textbf{1.33} & 1.23 & 1.13 & 1.16 & \btextbf{1.27} & 0.92\\
         & \textit{ratio $(\uparrow)$} & 1.57 & 1.72 & 1.51 & \textbf{1.88} & 1.82 & 1.62 & 1.24 & \btextbf{1.64}\\
\bottomrule
  \end{tabular}
    }
    \label{tab:distance_latent_features_mini}
\end{table}

Table~\ref{tab:greek_results_tot_paper} displays the classification outcomes from the various models. 
For each patch size, we initially present outcomes from the transformer models alongside our baseline, LOAN.
A pre-trained LOAN model is then fine-tuned using four separate contrastive methodologies: three using Eq.~\ref{eq:tri_loss} and one that employs Eq.~\ref{eq:scl_loss}.
The suffixes indicate the adopted training sampling strategies: \textit{LTL} denotes label sampling with triplet loss, \textit{HTL} represents historical sampling with triplet loss, and \textit{CTL} curriculum sampling with triplet loss. \textit{SCL} describes the model utilizing supervised contrastive loss. 
Finally, we present results for three of the four models trained across all epochs using the contrastive framework, as historical sampling is suboptimal in this context due to data constraints.

The results substantiate the performance enhancements facilitated by the CL approach in four scenarios. 
They show a significant performance improvement in the CL method that employs curriculum sampling, especially when compared with other CL outcomes.
Due to this specific sampling method, the model maintains its performance even with a reduced patch size of $5\times5$.
This indicates that the necessary FLOPS can be reduced from $168.2M$ (of the input $25\times25$) to $7.8M$ without impacting performance, or down to just $664.4k$ with a minor performance decrease by adopting a patch size $1\times1$.

As noted previously, we believe that the significant variability in dynamic features among same-class samples (within the context of CL) acts as a source of noise, hindering the model's ability to learn discriminative latent representations. 
This is substantiated by the findings in Table \ref{tab:distance_latent_features_mini}, which present an analysis of the models' latent spaces. 
We calculate the mean pairwise distance among normalized latent vectors from a subset of the original dataset\footnote{This subset is obtained by collecting all positive samples from the test dataset ($5635$ samples) and randomly selecting an equal amount of negative samples.}.
In the table, The \textit{Pos-Pos} and \textit{Neg-Neg} distances are denoted by the mean intra-class distance, while \textit{Pos-Neg} distances are indicated by the mean inter-class distance. 
We normalize the latent vectors before computing the distance. 
We also offer the ratio of inter-class to intra-class distances for a clearer comparison of different training methods.

Table~\ref{tab:distance_latent_features_mini} shows that historical sampling improves the structure of the latent space, resulting in greater distances between classes and consistent intra-class distances in various patch sizes. 
However, this enhancement does not translate directly to better performance in the classification task, where it remains comparable to other contrastive methods.
Finally, curriculum sampling does help to better shape the latent space, when the contrastive signal is adopted throughout the entire training, and also reach higher classification performance. 

\paragraph{Calabria Dataset.}

\begin{table}[ht]
  \caption{Overview of the aggregated metrics computed for the years 2017 and 2018 using the Calabria dataset. In this setting, patches are not centered on the target event; rather, wildfire occurrences may appear at any location within the patch. Each reported value corresponds to the mean over five independent trials. Class-wise results are provided in Appendix D of the Supplementary Material~\citep{loscudo2025}.}
  \centering
  \resizebox{.9\linewidth}{!}{
  \begin{tabular}{lc|c|c|c}
\multicolumn{5}{l}{\textbf{Aggregate}} \\
Model & \textit{Precision} & \textit{AUROC} & \textit{IoU} & \textit{F1}\\
\midrule
FWI & 0.67 \tiny{ $\pm$ 0.032} & 0.72 \tiny{ $\pm$0.002} & 0.50 \tiny{ $\pm$0.034} & 0.66  \tiny{ $\pm$0.030} \\
\midrule
TimeSformer & 0.83 \tiny{ $\pm$ 1e-02} & 0.91 \tiny{ $\pm$2e-03} & 0.70 \tiny{ $\pm$8e-03} & 0.83  \tiny{ $\pm$6e-03} \\
 SwinTransformer3D & 0.79 \tiny{ $\pm$ 7e-02} & 0.85 \tiny{ $\pm$3e-03} & 0.63 \tiny{ $\pm$3e-02} & 0.77  \tiny{ $\pm$3e-02} \\
 LOAN \small{(Baseline)} & 0.89 \tiny{ $\pm$ 3e-02} & 0.95 \tiny{ $\pm$2e-03} & 0.80 \tiny{ $\pm$8e-03} & 0.89  \tiny{ $\pm$5e-03} \\
  \cmidrule(r){1-5}
 LOAN+\small{LTL - ft} & 0.97 \tiny{ $\pm$ 4e-03} & 1.00 \tiny{ $\pm$2e-04} & 0.95 \tiny{ $\pm$2e-03} & 0.97  \tiny{ $\pm$9e-04} \\
 LOAN+\small{SCL - ft} & 0.96 \tiny{ $\pm$ 3e-02} & 0.99 \tiny{ $\pm$2e-04} & 0.92 \tiny{ $\pm$3e-03} & 0.96  \tiny{ $\pm$1e-03} \\
 LOAN+\small{HTL - ft (Ours)} & 0.73 \tiny{ $\pm$ 8e-02} & 0.77 \tiny{ $\pm$2e-03} & 0.54 \tiny{ $\pm$6e-02} & 0.70  \tiny{ $\pm$5e-02} \\
 LOAN+\small{CTL - ft (Ours)} & 0.85 \tiny{ $\pm$ 1e-02} & 0.92 \tiny{ $\pm$3e-03} & 0.74 \tiny{ $\pm$6e-03} & 0.85  \tiny{ $\pm$4e-03} \\
  \cmidrule(r){1-5}
 LOAN+\small{LTL - Full} & 0.97 \tiny{ $\pm$ 3e-02} & 0.99 \tiny{ $\pm$2e-04} & 0.94 \tiny{ $\pm$2e-03} & 0.97  \tiny{ $\pm$9e-04} \\
 LOAN+\small{SCL - Full} & \textbf{0.98} \tiny{ $\pm$ 1e-02} & \textbf{1.00} \tiny{ $\pm$5e-05} & \textbf{0.97} \tiny{ $\pm$8e-04} & \textbf{0.98}  \tiny{ $\pm$4e-04} \\
 LOAN+\small{CTL - Full (Ours)} & \textbf{0.98} \tiny{ $\pm$ 2e-02} & 0.99 \tiny{ $\pm$2e-04} & 0.95 \tiny{ $\pm$2e-03} & \textbf{0.98}  \tiny{ $\pm$9e-04} \\
\bottomrule
  \end{tabular}
  }
\label{tab:calabria_results_tot}
\end{table}

The classification task outcomes derived from the Calabria dataset are summarized in Table~\ref{tab:calabria_results_tot}.
Within this new application scenario, our sampling strategies exhibit reduced efficacy during the fine-tuning phase, whereas curriculum sampling matches the SCL method's performance when applied for the entire training.
This situation is justified by the difference in the dynamic features of the two datasets. 
The Figures~\ref{fig:box_dyn_greece} and~\ref{fig:box_dyn_cal} in Section~\ref{sec:dataset} demonstrate that the Calabria dataset exhibits greater feature regularity between the two classes.
Consequently, our sampling method yields a diminished benefit as label-based sampling already guarantees sample similarity.

However, despite being more challenging, the models utilizing a CL approach achieve consistent gains over the baseline due to the higher resolution of the data.
In the full-training approach, all evaluated models yield similar results with only minimal variation.

\section{Conclusions}\label{sec:concl}

This study presents an innovative methodology to enhance contrastive learning (CL) for wildfire risk prediction through curriculum data, improving model robustness and accuracy. 
By integrating similarity-based perspectives into the CL framework, this approach addresses limitations in current methods, providing a more effective solution. 
To our knowledge, this is the first systematic analysis of CL in predicting wildfire risk.

The experimental results in Section~\ref{sec:results} confirm our methodology’s effectiveness. Across diverse patch sizes, our model consistently outperformed both the baseline and conventional CL models. 
In more complex scenarios, where wildfire events may occur anywhere within the patch, the model demonstrated strong generalization and classification accuracy, reinforcing its robustness. 
Integrating curriculum sampling into the CL framework improved model performance, reliability, and robustness across various conditions while reducing computational costs without sacrificing accuracy.

Although CL is a consolidate approach, our proposed morphology-aware curriculum CL paves the way for advancements by enabling future research to refine it through enhanced training sample selection based on similarity measures in autoencoder-derived latent spaces. 
Utilizing autoencoders to generate latent representations could allow the identification and selection of more representative and diverse training samples, likely improving model generalizability. 
Future work could also explore self-supervised techniques that leverage intrinsic temporal and spatial data patterns to generate pseudo-labels, facilitating learning of recurring structures in historical data.

In practical applications, organizations and institutions whose responsibilities or operational mandates are directly or indirectly concerned with the prevention, management, or mitigation of wildfire hazards may readily integrate the proposed approach into their existing systems.
Adopting this approach could provide high-accuracy predictions and, under specific circumstances, also reduce computational cost.




\begin{ack}
\small{This work was funded by the Next Generation EU - Italian NRRP, Mission 4, Component 2, Investment 1.5, call for the creation and strengthening of ``Innovation Ecosystems'', building ``Territorial R\&D Leaders'' (Directorial Decree n. 2021/3277) - project Tech4You - Technologies for climate change adaptation and quality of life improvement, n. ECS0000009. This work reflects only the authors’ views and opinions, neither the Ministry for University and Research nor the European Commission can be considered responsible for them.}
\end{ack}



\bibliography{m577}
\appendix
\begin{table*}[ht]
     \caption{Average Difference over Dynamic Features from the Calabria Dataset computed using triplets chosen solely based on label data, triplets selected through our historical methodology, and finally through our curriculum methodology. 
    }
    \centering
    \resizebox{.9\linewidth}{!}{
    \begin{tabular}{rccc|ccc|ccc}
    & \multicolumn{3}{c}{\textit{Avg. Anchor-Positive Diff $(\downarrow)$}} & \multicolumn{3}{c}{Avg. Anchor-Negative Diff $(\uparrow)$}  & \multicolumn{3}{c}{\textit{ratio $(\uparrow)$}}\\
    Feature & \textit{Label} & \textit{Historical} & \textit{Curriculum}  & \textit{Label} & \textit{Historical }  & \textit{Curriculum} & \textit{Label} & \textit{Hist.}  & \textit{Curr.} \\
    \midrule
Pprecmap & 0.06 \tiny{ $\pm$ 7e-02} & 0.05 \tiny{ $\pm$ 7e-02} & 0.06 \tiny{ $\pm$ 6e-02} & 0.08 \tiny{ $\pm$ 8e-02} &0.11 \tiny{ $\pm$ 8e-02} &0.05 \tiny{ $\pm$ 6e-02} &1.3 & \textbf{2.1} & 0.9 \\
RHmap & 0.35 \tiny{ $\pm$ 1e-01} & 0.27 \tiny{ $\pm$ 1e-01} & 0.28 \tiny{ $\pm$ 1e-01} & 0.35 \tiny{ $\pm$ 1e-01} &0.32 \tiny{ $\pm$ 1e-01} &0.26 \tiny{ $\pm$ 1e-01} &1.0 & \textbf{1.2} & 0.9 \\
Tmap & 0.33 \tiny{ $\pm$ 1e-01} & 0.23 \tiny{ $\pm$ 1e-01} & 0.27 \tiny{ $\pm$ 1e-01} & 0.37 \tiny{ $\pm$ 1e-01} &0.31 \tiny{ $\pm$ 1e-01} &0.30 \tiny{ $\pm$ 1e-01} &1.1 & \textbf{1.4} & 1.1 \\
Vmap & 0.19 \tiny{ $\pm$ 8e-02} & 0.18 \tiny{ $\pm$ 1e-01} & 0.14 \tiny{ $\pm$ 7e-02} & 0.19 \tiny{ $\pm$ 8e-02} &0.15 \tiny{ $\pm$ 8e-02} &0.16 \tiny{ $\pm$ 8e-02} &1.0 & 0.8 & \textbf{1.2} \\
dwr & 0.13 \tiny{ $\pm$ 7e-02} & 0.16 \tiny{ $\pm$ 9e-02} & 0.10 \tiny{ $\pm$ 7e-02} & 0.14 \tiny{ $\pm$ 8e-02} &0.14 \tiny{ $\pm$ 1e-01} &0.11 \tiny{ $\pm$ 6e-02} &\textbf{1.1} & 0.9 & \textbf{1.1} \\
Fwi & 0.33 \tiny{ $\pm$ 1e-01} & 0.36 \tiny{ $\pm$ 1e-01} & 0.28 \tiny{ $\pm$ 1e-01} & 0.37 \tiny{ $\pm$ 1e-01} &0.38 \tiny{ $\pm$ 1e-01} &0.29 \tiny{ $\pm$ 1e-01} &\textbf{1.1} & \textbf{1.1} & \textbf{1.1} \\
Isi & 0.22 \tiny{ $\pm$ 8e-02} & 0.27 \tiny{ $\pm$ 1e-01} & 0.19 \tiny{ $\pm$ 9e-02} & 0.24 \tiny{ $\pm$ 9e-02} &0.25 \tiny{ $\pm$ 1e-01} &0.20 \tiny{ $\pm$ 9e-02} &\textbf{1.1} & 0.9 & \textbf{1.1} \\
Bui & 0.25 \tiny{ $\pm$ 1e-01} & 0.31 \tiny{ $\pm$ 1e-01} & 0.20 \tiny{ $\pm$ 1e-01} & 0.28 \tiny{ $\pm$ 1e-01} &0.30 \tiny{ $\pm$ 2e-01} &0.22 \tiny{ $\pm$ 9e-02} &\textbf{1.1} & 1.0 & \textbf{1.1} \\
Dmc & 0.40 \tiny{ $\pm$ 1e-01} & 0.38 \tiny{ $\pm$ 1e-01} & 0.34 \tiny{ $\pm$ 1e-01} & 0.43 \tiny{ $\pm$ 1e-01} &0.38 \tiny{ $\pm$ 2e-01} &0.35 \tiny{ $\pm$ 1e-01} &\textbf{1.1} & 1.0 & 1.0 \\
Fmc & 0.20 \tiny{ $\pm$ 1e-01} & 0.26 \tiny{ $\pm$ 1e-01} & 0.17 \tiny{ $\pm$ 1e-01} & 0.23 \tiny{ $\pm$ 1e-01} &0.25 \tiny{ $\pm$ 1e-01} &0.19 \tiny{ $\pm$ 9e-02} &\textbf{1.1} & 1.0 & \textbf{1.1} \\
\bottomrule
    \end{tabular}
    }
    \label{tab:delta_features_calabria}
\end{table*}

\section{Constrastive samplings}

\begin{figure}[ht]
\centering
\includegraphics[width=8.cm]{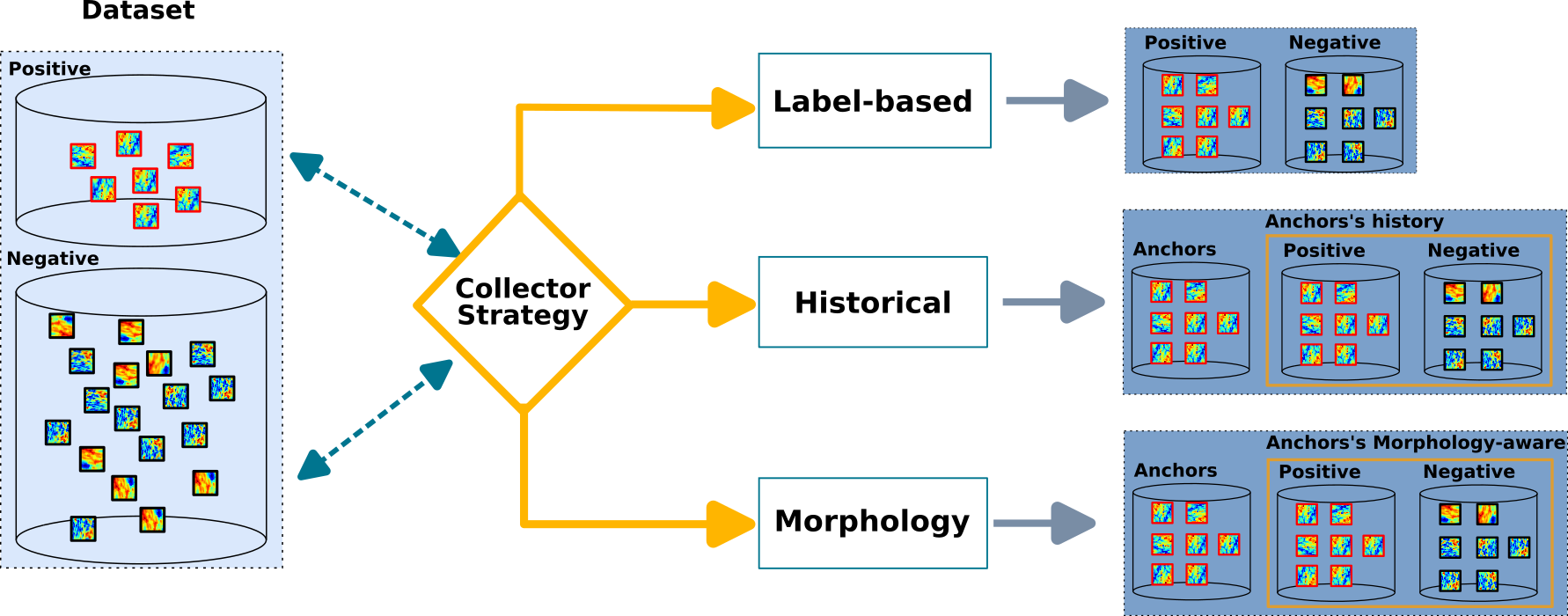}
\caption{Illustration of the three sampling strategies used in this work.}
\label{fig:sampling_strategies}
\end{figure}

Three distinct sampling strategies are employed to enable the different contrastive learning approaches proposed in this study.

The traditional label-based (CL) method utilizes the entire dataset. For each anchor item, positive and negative samples are selected based solely on label information. However, this strategy may produce ambiguous training signals, as samples sharing the same label can exhibit markedly different dynamic behaviors due to the heterogeneous nature of the regions they represent.

The historical sampling strategy addresses this issue by computing positive and negative sample sets for each patch based exclusively on its historical data. During training, triplets are then constructed using these precomputed sets, thereby capturing temporal consistency in local wildfire patterns.

Finally, the morphology-aware approach generalizes the historical method by relaxing its constraints—specifically, the limitation to patches with positive occurrences. Instead, it defines positive and negative sets based on morphological similarity across all patches in the dataset, thereby enabling the model to leverage structural patterns inherent in the terrain.

\section{Training protocols}

\begin{figure}[ht]
\centering
\includegraphics[width=7.cm]{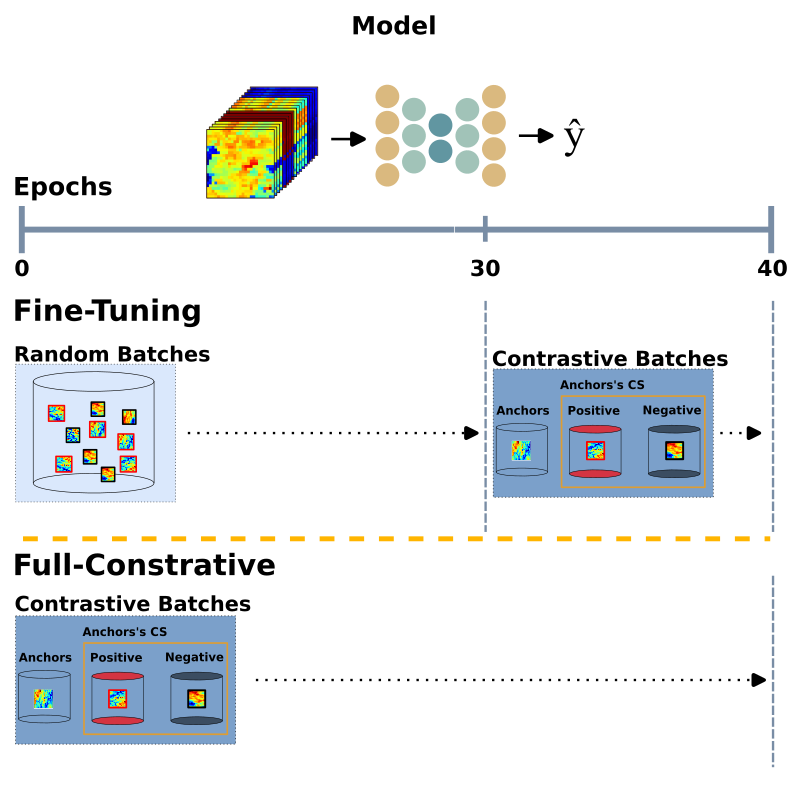}
\caption{Illustration of the two training strategies used in this work.}
\label{fig:training_protocols}
\end{figure}

In our experimental evaluation, we examine two distinct training protocols.
In the first protocol, CL is used exclusively as a fine-tuning stage, following an initial standard training phase conducted on the full dataset. This approach ensures that the model is exposed to sufficient data to support robust generalization before introducing the contrastive objective.
In the second protocol, the entire training process is carried out under the CL framework from the outset.
For both protocols, we systematically assess the performance of the various CL strategies implemented, in order to evaluate their relative effectiveness under different training regimes.

\section{Effect of the sampling strategies on the Calabria Dataset}
\label{sec:delta_features_calabria}

Table~\ref{tab:delta_features_calabria} reports the average normalized difference for dynamic features, calculated using triplet-based comparisons.
For each anchor patch, we randomly choose ten positive and ten negative samples based on label-based sampling. Similarly, for historical and curriculum sampling, we construct ten triplets using pre-computed maps.

The results show that a similar ratio (the mean distance between the dynamic features of the anchor and the negative samples and the anchor and the positive samples, $\delta(x_d^a,x_d^n)/\delta(x_d^a,x_d^p)$) across different sampling strategies. 
This consistency arises because, unlike the Greece scenario where data distribution differences are notable, the variation among classes in dynamic features is subtler. 
This factor also accounts for the comparable performance of label and curriculum-based CL methods.

\newpage

\section{Classification task results by class}

In Tables~\ref{tab:greek_results_by_class_bg},~\ref{tab:greek_results_by_class_wf}, and~\ref{tab:calabria_results_by_class}, we present the detailed outcomes (by class) of the classification analysis conducted on the datasets from Greece and Calabria.

\begin{table}[!ht]
\caption{Background Results computed over the years 2020 and 2021 of the FireCube Dataset. Each reported value represents the mean of five independent trials.}
  \centering
  \resizebox{1.\linewidth}{!}{
  \begin{tabular}{clc|c|c|c}
PS & Model & \textit{Precision} & \textit{Accuracy} & \textit{IoU} & \textit{F1}\\
    \midrule
        \multirow{10}{*}{\rotatebox[origin=c]{90}{$1\times1$}} & 
    TimeSformer & 0.88 \tiny{ $\pm$6e-03} & 0.92 \tiny{ $\pm$6e-03} & 0.82 \tiny{ $\pm$3e-03} & 0.90 \tiny{ $\pm$2e-03} \\
     & SwinTransformer3D & 0.87 \tiny{ $\pm$1e-02} & 0.91 \tiny{ $\pm$1e-02} & 0.80 \tiny{ $\pm$5e-03} & 0.89 \tiny{ $\pm$3e-03} \\
     & LOAN \small{(Baseline)} & 0.85 \tiny{ $\pm$2e-03} & 0.96 \tiny{ $\pm$1e-03} & 0.82 \tiny{ $\pm$2e-03} & 0.90 \tiny{ $\pm$1e-03} \\
      \cmidrule(r){2-6}
     & LOAN+\small{HTL - ft (Ours)} & 0.89 \tiny{ $\pm$3e-03} & 0.91 \tiny{ $\pm$8e-03} & 0.83 \tiny{ $\pm$5e-03} & 0.90 \tiny{ $\pm$3e-03} \\
     & LOAN+\small{LTL - ft} & 0.88 \tiny{ $\pm$2e-02} & 0.95 \tiny{ $\pm$9e-03} & 0.84 \tiny{ $\pm$8e-03} & 0.91 \tiny{ $\pm$5e-03} \\
     & LOAN+\small{SCL - ft} & 0.86 \tiny{ $\pm$2e-02} & 0.95 \tiny{ $\pm$8e-03} & 0.82 \tiny{ $\pm$2e-02} & 0.90 \tiny{ $\pm$9e-03} \\
      & LOAN+\small{CTL - ft (Ours)} & 0.92 \tiny{ $\pm$2e-03} & 0.95 \tiny{ $\pm$5e-03} & 0.87 \tiny{ $\pm$5e-03} & 0.93 \tiny{ $\pm$3e-03} \\
       \cmidrule(r){2-6}
     & LOAN+\small{LTL - full} & 0.89 \tiny{ $\pm$6e-03} & 0.94 \tiny{ $\pm$5e-03} & 0.84 \tiny{ $\pm$4e-03} & 0.91 \tiny{ $\pm$3e-03} \\
     & LOAN+\small{SCL - full} & 0.88 \tiny{ $\pm$1e-02} & 0.95 \tiny{ $\pm$7e-03} & 0.84 \tiny{ $\pm$1e-02} & 0.91 \tiny{ $\pm$6e-03} \\
      & LOAN+\small{CTL - full (Ours)} & 0.91 \tiny{ $\pm$9e-03} & 0.96 \tiny{ $\pm$1e-02} & 0.88 \tiny{ $\pm$3e-03} & \textbf{0.94} \tiny{ $\pm$2e-03} \\
        \midrule
        \multirow{10}{*}{\rotatebox[origin=c]{90}{$5\times5$}} & 
        TimeSformer & 0.84 \tiny{ $\pm$6e-02} & 0.93 \tiny{ $\pm$1e-02} & 0.79 \tiny{ $\pm$4e-02} & 0.88 \tiny{ $\pm$3e-02} \\
     & SwinTransformer3D & 0.88 \tiny{ $\pm$7e-03} & 0.94 \tiny{ $\pm$1e-02} & 0.83 \tiny{ $\pm$6e-03} & 0.91 \tiny{ $\pm$4e-03} \\
     & LOAN \small{(Baseline)} & 0.82 \tiny{ $\pm$3e-02} & 0.97 \tiny{ $\pm$2e-03} & 0.80 \tiny{ $\pm$3e-02} & 0.89 \tiny{ $\pm$2e-02} \\
      \cmidrule(r){2-6}
     & LOAN+\small{HTL - ft (Ours)} & 0.89 \tiny{ $\pm$6e-03} & 0.94 \tiny{ $\pm$1e-02} & 0.84 \tiny{ $\pm$4e-03} & 0.91 \tiny{ $\pm$3e-03} \\
     & LOAN+\small{LTL - ft} & 0.85 \tiny{ $\pm$5e-02} & 0.95 \tiny{ $\pm$1e-02} & 0.81 \tiny{ $\pm$4e-02} & 0.90 \tiny{ $\pm$2e-02} \\
     & LOAN+\small{SCL - ft} & 0.86 \tiny{ $\pm$4e-02} & 0.96 \tiny{ $\pm$1e-02} & 0.83 \tiny{ $\pm$3e-02} & 0.91 \tiny{ $\pm$2e-02} \\
      & LOAN+\small{CTL - ft (Ours)} & 0.93 \tiny{ $\pm$6e-03} & 0.96 \tiny{ $\pm$7e-03} & 0.89 \tiny{ $\pm$3e-03} & 0.94 \tiny{ $\pm$1e-03} \\
       \cmidrule(r){2-6}
     & LOAN+\small{LTL - full} & 0.84 \tiny{ $\pm$5e-02} & 0.95 \tiny{ $\pm$1e-02} & 0.81 \tiny{ $\pm$4e-02} & 0.89 \tiny{ $\pm$2e-02} \\
     & LOAN+\small{SCL - full} & 0.87 \tiny{ $\pm$4e-02} & 0.96 \tiny{ $\pm$5e-03} & 0.84 \tiny{ $\pm$3e-02} & 0.91 \tiny{ $\pm$2e-02} \\
      & LOAN+\small{CTL - full (Ours)} & 0.93 \tiny{ $\pm$2e-02} & 0.97 \tiny{ $\pm$1e-02} & 0.90 \tiny{ $\pm$8e-03} & \textbf{0.95} \tiny{ $\pm$4e-03} \\
                \midrule
        \multirow{10}{*}{\rotatebox[origin=c]{90}{$15\times15$}} & 
        TimeSformer & 0.85 \tiny{ $\pm$4e-02} & 0.93 \tiny{ $\pm$2e-02} & 0.80 \tiny{ $\pm$3e-02} & 0.89 \tiny{ $\pm$2e-02} \\
     & SwinTransformer3D & 0.88 \tiny{ $\pm$7e-03} & 0.94 \tiny{ $\pm$1e-02} & 0.83 \tiny{ $\pm$6e-03} & 0.91 \tiny{ $\pm$4e-03} \\
     & LOAN \small{(Baseline)} & 0.82 \tiny{ $\pm$3e-02} & 0.97 \tiny{ $\pm$2e-03} & 0.80 \tiny{ $\pm$3e-02} & 0.89 \tiny{ $\pm$2e-02} \\
      \cmidrule(r){2-6}
     & LOAN+\small{HTL - ft (Ours)} & 0.89 \tiny{ $\pm$7e-03} & 0.94 \tiny{ $\pm$9e-03} & 0.84 \tiny{ $\pm$4e-03} & 0.91 \tiny{ $\pm$2e-03} \\
     & LOAN+\small{LTL - ft} & 0.84 \tiny{ $\pm$6e-02} & 0.95 \tiny{ $\pm$2e-02} & 0.81 \tiny{ $\pm$4e-02} & 0.89 \tiny{ $\pm$3e-02} \\
     & LOAN+\small{SCL - ft} & 0.85 \tiny{ $\pm$6e-02} & 0.96 \tiny{ $\pm$2e-02} & 0.81 \tiny{ $\pm$5e-02} & 0.90 \tiny{ $\pm$3e-02} \\
      & LOAN+\small{CTL - ft (Ours)} & 0.93 \tiny{ $\pm$6e-03} & 0.96 \tiny{ $\pm$7e-03} & 0.90 \tiny{ $\pm$3e-03} & 0.94 \tiny{ $\pm$2e-03} \\
       \cmidrule(r){2-6}
     & LOAN+\small{LTL - full} & 0.85 \tiny{ $\pm$4e-02} & 0.96 \tiny{ $\pm$9e-03} & 0.81 \tiny{ $\pm$3e-02} & 0.90 \tiny{ $\pm$2e-02} \\
     & LOAN+\small{SCL - full} & 0.86 \tiny{ $\pm$5e-02} & 0.96 \tiny{ $\pm$1e-02} & 0.83 \tiny{ $\pm$4e-02} & 0.91 \tiny{ $\pm$2e-02} \\
      & LOAN+\small{CTL - full (Ours)} & 0.92 \tiny{ $\pm$2e-02} & 0.97 \tiny{ $\pm$1e-02} & 0.90 \tiny{ $\pm$1e-02} & \textbf{0.95} \tiny{ $\pm$5e-03} \\
        \midrule
        \multirow{10}{*}{\rotatebox[origin=c]{90}{$25\times25$}} & 
        TimeSformer & 0.85 \tiny{ $\pm$4e-02} & 0.93 \tiny{ $\pm$2e-02} & 0.79 \tiny{ $\pm$2e-02} & 0.88 \tiny{ $\pm$1e-02} \\
     & SwinTransformer3D & 0.88 \tiny{ $\pm$3e-03} & 0.94 \tiny{ $\pm$8e-03} & 0.83 \tiny{ $\pm$5e-03} & 0.91 \tiny{ $\pm$3e-03} \\
     & LOAN \small{(Baseline)} & 0.86 \tiny{ $\pm$3e-02} & 0.96 \tiny{ $\pm$7e-03} & 0.84 \tiny{ $\pm$2e-02} & 0.91 \tiny{ $\pm$1e-02} \\
      \cmidrule(r){2-6}
     & LOAN+\small{HTL - ft (Ours)} & 0.88 \tiny{ $\pm$1e-02} & 0.95 \tiny{ $\pm$7e-03} & 0.84 \tiny{ $\pm$6e-03} & 0.91 \tiny{ $\pm$3e-03} \\
     & LOAN+\small{LTL - ft} & 0.88 \tiny{ $\pm$4e-02} & 0.96 \tiny{ $\pm$3e-02} & 0.84 \tiny{ $\pm$1e-02} & 0.91 \tiny{ $\pm$9e-03} \\
     & LOAN+\small{SCL - ft} & 0.89 \tiny{ $\pm$6e-03} & 0.95 \tiny{ $\pm$6e-03} & 0.85 \tiny{ $\pm$8e-03} & 0.92 \tiny{ $\pm$5e-03} \\
      & LOAN+\small{CTL - ft (Ours)} & 0.92 \tiny{ $\pm$9e-03} & 0.98 \tiny{ $\pm$6e-03} & 0.90 \tiny{ $\pm$4e-03} & \textbf{0.95} \tiny{ $\pm$2e-03} \\
       \cmidrule(r){2-6}
     & LOAN+\small{LTL - full} & 0.85 \tiny{ $\pm$6e-02} & 0.97 \tiny{ $\pm$1e-02} & 0.83 \tiny{ $\pm$5e-02} & 0.90 \tiny{ $\pm$3e-02} \\
     & LOAN+\small{SCL - full} & 0.89 \tiny{ $\pm$1e-02} & 0.96 \tiny{ $\pm$7e-03} & 0.86 \tiny{ $\pm$1e-02} & 0.92 \tiny{ $\pm$8e-03} \\
      & LOAN+\small{CTL - full (Ours)} & 0.93 \tiny{ $\pm$2e-02} & 0.98 \tiny{ $\pm$1e-02} & 0.91 \tiny{ $\pm$2e-02} & \textbf{0.95} \tiny{ $\pm$9e-03} \\
    \bottomrule
\end{tabular}
  }
\label{tab:greek_results_by_class_bg}
\end{table}

\begin{table}[!ht]
\caption{Wildfire results computed over the years 2020 and 2021 of the FireCube Dataset. Each reported value represents the mean of five independent trials.}
  \centering
  \resizebox{1.\linewidth}{!}{
  \begin{tabular}{clc|c|c|c}
    PS & Model & \textit{Precision} & \textit{Accuracy} & \textit{IoU} & \textit{F1}\\
        \midrule
    
    \multirow{10}{*}{\rotatebox[origin=c]{90}{$1\times1$}} & 
TimeSformer & 0.92 \tiny{ $\pm$5e-03} & 0.88 \tiny{ $\pm$7e-03} & 0.81 \tiny{ $\pm$4e-03} & 0.90 \tiny{ $\pm$2e-03} \\
 & SwinTransformer3D & 0.91 \tiny{ $\pm$1e-02} & 0.86 \tiny{ $\pm$1e-02} & 0.79 \tiny{ $\pm$7e-03} & 0.88 \tiny{ $\pm$4e-03} \\
 & LOAN \small{(Baseline)} & 0.95 \tiny{ $\pm$1e-03} & 0.83 \tiny{ $\pm$3e-03} & 0.80 \tiny{ $\pm$3e-03} & 0.89 \tiny{ $\pm$2e-03} \\
  \cmidrule(r){2-6}
 & LOAN+\small{HTL - ft (Ours)} & 0.91 \tiny{ $\pm$7e-03} & 0.89 \tiny{ $\pm$4e-03} & 0.82 \tiny{ $\pm$4e-03} & 0.90 \tiny{ $\pm$3e-03} \\
 & LOAN+\small{LTL - ft} & 0.94 \tiny{ $\pm$9e-03} & 0.86 \tiny{ $\pm$2e-02} & 0.82 \tiny{ $\pm$1e-02} & 0.90 \tiny{ $\pm$8e-03} \\
 & LOAN+\small{SCL - ft} & 0.94 \tiny{ $\pm$7e-03} & 0.84 \tiny{ $\pm$3e-02} & 0.80 \tiny{ $\pm$2e-02} & 0.89 \tiny{ $\pm$1e-02} \\
  & LOAN+\small{CTL - ft (Ours)} & 0.94 \tiny{ $\pm$5e-03} & 0.91 \tiny{ $\pm$2e-03} & 0.87 \tiny{ $\pm$4e-03} & \textbf{0.93} \tiny{ $\pm$3e-03} \\
   \cmidrule(r){2-6}
 & LOAN+\small{LTL - full} & 0.93 \tiny{ $\pm$5e-03} & 0.89 \tiny{ $\pm$8e-03} & 0.84 \tiny{ $\pm$5e-03} & 0.91 \tiny{ $\pm$3e-03} \\
 & LOAN+\small{SCL - full} & 0.94 \tiny{ $\pm$6e-03} & 0.87 \tiny{ $\pm$2e-02} & 0.83 \tiny{ $\pm$1e-02} & 0.90 \tiny{ $\pm$8e-03} \\
  & LOAN+\small{CTL - full (Ours)} & 0.96 \tiny{ $\pm$1e-02} & 0.91 \tiny{ $\pm$1e-02} & 0.87 \tiny{ $\pm$4e-03} & \textbf{0.93} \tiny{ $\pm$2e-03} \\
    \midrule
    \multirow{10}{*}{\rotatebox[origin=c]{90}{$5\times5$}} &
TimeSformer & 0.92 \tiny{ $\pm$9e-03} & 0.81 \tiny{ $\pm$8e-02} & 0.76 \tiny{ $\pm$7e-02} & 0.86 \tiny{ $\pm$4e-02} \\
 & SwinTransformer3D & 0.93 \tiny{ $\pm$1e-02} & 0.88 \tiny{ $\pm$1e-02} & 0.82 \tiny{ $\pm$5e-03} & 0.90 \tiny{ $\pm$3e-03} \\
 & LOAN \small{(Baseline)} & 0.96 \tiny{ $\pm$3e-03} & 0.78 \tiny{ $\pm$4e-02} & 0.76 \tiny{ $\pm$4e-02} & 0.86 \tiny{ $\pm$3e-02} \\
  \cmidrule(r){2-6}
 & LOAN+\small{HTL - ft (Ours)} & 0.93 \tiny{ $\pm$9e-03} & 0.89 \tiny{ $\pm$8e-03} & 0.83 \tiny{ $\pm$3e-03} & 0.91 \tiny{ $\pm$2e-03} \\
 & LOAN+\small{LTL - ft} & 0.95 \tiny{ $\pm$9e-03} & 0.83 \tiny{ $\pm$7e-02} & 0.79 \tiny{ $\pm$6e-02} & 0.88 \tiny{ $\pm$4e-02} \\
 & LOAN+\small{SCL - ft} & 0.96 \tiny{ $\pm$1e-02} & 0.84 \tiny{ $\pm$6e-02} & 0.81 \tiny{ $\pm$5e-02} & 0.89 \tiny{ $\pm$3e-02} \\
  & LOAN+\small{CTL - ft (Ours)} & 0.96 \tiny{ $\pm$7e-03} & 0.92 \tiny{ $\pm$7e-03} & 0.89 \tiny{ $\pm$3e-03} & 0.94 \tiny{ $\pm$1e-03} \\
   \cmidrule(r){2-6}
 & LOAN+\small{LTL - full} & 0.95 \tiny{ $\pm$1e-02} & 0.81 \tiny{ $\pm$8e-02} & 0.78 \tiny{ $\pm$6e-02} & 0.87 \tiny{ $\pm$4e-02} \\
 & LOAN+\small{SCL - full} & 0.96 \tiny{ $\pm$4e-03} & 0.86 \tiny{ $\pm$5e-02} & 0.82 \tiny{ $\pm$4e-02} & 0.90 \tiny{ $\pm$3e-02} \\
  & LOAN+\small{CTL - full (Ours)} & 0.97 \tiny{ $\pm$1e-02} & 0.93 \tiny{ $\pm$2e-02} & 0.90 \tiny{ $\pm$1e-02} & \textbf{0.95} \tiny{ $\pm$6e-03} \\
    \midrule
    \multirow{10}{*}{\rotatebox[origin=c]{90}{$15\times 15$}} & 
TimeSformer & 0.92 \tiny{ $\pm$1e-02} & 0.83 \tiny{ $\pm$6e-02} & 0.78 \tiny{ $\pm$4e-02} & 0.87 \tiny{ $\pm$3e-02} \\
 & SwinTransformer3D & 0.93 \tiny{ $\pm$1e-02} & 0.88 \tiny{ $\pm$1e-02} & 0.82 \tiny{ $\pm$5e-03} & 0.90 \tiny{ $\pm$3e-03} \\
 & LOAN \small{(Baseline)} & 0.96 \tiny{ $\pm$3e-03} & 0.78 \tiny{ $\pm$4e-02} & 0.76 \tiny{ $\pm$4e-02} & 0.86 \tiny{ $\pm$3e-02} \\
  \cmidrule(r){2-6}
 & LOAN+\small{HTL - ft (Ours)} & 0.94 \tiny{ $\pm$9e-03} & 0.88 \tiny{ $\pm$9e-03} & 0.83 \tiny{ $\pm$4e-03} & 0.91 \tiny{ $\pm$2e-03} \\
 & LOAN+\small{LTL - ft} & 0.94 \tiny{ $\pm$2e-02} & 0.82 \tiny{ $\pm$8e-02} & 0.78 \tiny{ $\pm$7e-02} & 0.87 \tiny{ $\pm$4e-02} \\
 & LOAN+\small{SCL - ft} & 0.95 \tiny{ $\pm$1e-02} & 0.82 \tiny{ $\pm$9e-02} & 0.78 \tiny{ $\pm$8e-02} & 0.88 \tiny{ $\pm$5e-02} \\
  & LOAN+\small{CTL - ft (Ours)} & 0.96 \tiny{ $\pm$7e-03} & 0.92 \tiny{ $\pm$7e-03} & 0.89 \tiny{ $\pm$3e-03} & 0.94 \tiny{ $\pm$2e-03} \\
   \cmidrule(r){2-6}
 & LOAN+\small{LTL - full} & 0.95 \tiny{ $\pm$8e-03} & 0.82 \tiny{ $\pm$6e-02} & 0.79 \tiny{ $\pm$5e-02} & 0.88 \tiny{ $\pm$3e-02} \\
 & LOAN+\small{SCL - full} & 0.96 \tiny{ $\pm$1e-02} & 0.84 \tiny{ $\pm$6e-02} & 0.81 \tiny{ $\pm$5e-02} & 0.90 \tiny{ $\pm$3e-02} \\
  & LOAN+\small{CTL - full (Ours)} & 0.97 \tiny{ $\pm$1e-02} & 0.92 \tiny{ $\pm$2e-02} & 0.89 \tiny{ $\pm$1e-02} & \textbf{0.94} \tiny{ $\pm$7e-03} \\
\midrule
    \multirow{10}{*}{\rotatebox[origin=c]{90}{$25\hspace{-0.6mm}\times\hspace{-0.6mm}25$}} & 
TimeSformer & 0.92 \tiny{ $\pm$2e-02} & 0.83 \tiny{ $\pm$5e-02} & 0.77 \tiny{ $\pm$3e-02} & 0.87 \tiny{ $\pm$2e-02} \\
 & SwinTransformer3D & 0.94 \tiny{ $\pm$7e-03} & 0.87 \tiny{ $\pm$5e-03} & 0.82 \tiny{ $\pm$4e-03} & 0.90 \tiny{ $\pm$2e-03} \\
 & LOAN \small{(Baseline)} & 0.96 \tiny{ $\pm$7e-03} & 0.85 \tiny{ $\pm$3e-02} & 0.82 \tiny{ $\pm$3e-02} & 0.90 \tiny{ $\pm$2e-02} \\
  \cmidrule(r){2-6}
 & LOAN+\small{HTL - ft (Ours)} & 0.94 \tiny{ $\pm$7e-03} & 0.88 \tiny{ $\pm$2e-02} & 0.83 \tiny{ $\pm$9e-03} & 0.91 \tiny{ $\pm$6e-03} \\
 & LOAN+\small{LTL - ft} & 0.95 \tiny{ $\pm$3e-02} & 0.87 \tiny{ $\pm$5e-02} & 0.83 \tiny{ $\pm$3e-02} & 0.91 \tiny{ $\pm$1e-02} \\
 & LOAN+\small{SCL - ft} & 0.95 \tiny{ $\pm$7e-03} & 0.88 \tiny{ $\pm$7e-03} & 0.84 \tiny{ $\pm$9e-03} & 0.91 \tiny{ $\pm$5e-03} \\
  & LOAN+\small{CTL - ft (Ours)} & 0.98 \tiny{ $\pm$6e-03} & 0.91 \tiny{ $\pm$1e-02} & 0.89 \tiny{ $\pm$6e-03} & 0.94 \tiny{ $\pm$3e-03} \\
   \cmidrule(r){2-6}
 & LOAN+\small{LTL - full} & 0.97 \tiny{ $\pm$1e-02} & 0.82 \tiny{ $\pm$8e-02} & 0.80 \tiny{ $\pm$7e-02} & 0.89 \tiny{ $\pm$5e-02} \\
 & LOAN+\small{SCL - full} & 0.96 \tiny{ $\pm$7e-03} & 0.88 \tiny{ $\pm$2e-02} & 0.84 \tiny{ $\pm$2e-02} & 0.91 \tiny{ $\pm$1e-02} \\
  & LOAN+\small{CTL - full (Ours)} & 0.98 \tiny{ $\pm$9e-03} & 0.92 \tiny{ $\pm$3e-02} & 0.90 \tiny{ $\pm$2e-02} & \textbf{0.95} \tiny{ $\pm$1e-02} \\
    \bottomrule
\end{tabular}
  }
\label{tab:greek_results_by_class_wf}
\end{table}

\begin{table}[!ht]
  \caption{Overview of metrics calculated for the years 2017 and 2018 using the Calabria Dataset. In this case, patches are not centered on the target event; instead, the wildfire event may occur at any location within the patch. Each value reported represents the mean of five independent trials.}
  \centering
  \resizebox{1.\linewidth}{!}{
  \begin{tabular}{lc|c|c|c}
  \multicolumn{5}{l}{\textbf{Background}}  \\
Model & \textit{Precision} & \textit{Accuracy} & \textit{IoU} & \textit{F1}\\
\midrule
FWI & 0.64 \tiny{ $\pm$0.00} & 0.76 \tiny{ $\pm$0.00} & 0.53 \tiny{ $\pm$0.00} & 0.69 \tiny{ $\pm$0.00}  \\
\midrule
TimeSformer & 0.84 \tiny{ $\pm$2e-03} & 0.81 \tiny{ $\pm$9e-03} & 0.70 \tiny{ $\pm$8e-03} & 0.82 \tiny{ $\pm$5e-03} \\
 SwinTransformer3D & 0.85 \tiny{ $\pm$1e-03} & 0.66 \tiny{ $\pm$6e-03} & 0.59 \tiny{ $\pm$5e-03} & 0.74 \tiny{ $\pm$4e-03} \\
 LOAN \small{(Baseline)} & 0.86 \tiny{ $\pm$9e-04} & 0.92 \tiny{ $\pm$7e-03} & 0.80 \tiny{ $\pm$6e-03} & 0.89 \tiny{ $\pm$4e-03} \\
  \cmidrule(r){1-5}
 LOAN+\small{LTL - ft} & 0.98 \tiny{ $\pm$4e-05} & 0.97 \tiny{ $\pm$2e-03} & 0.95 \tiny{ $\pm$2e-03} & 0.97 \tiny{ $\pm$1e-03} \\
 LOAN+\small{SCL - ft} & 0.93 \tiny{ $\pm$1e-04} & 0.99 \tiny{ $\pm$2e-03} & 0.93 \tiny{ $\pm$1e-03} & 0.96 \tiny{ $\pm$8e-04} \\
 LOAN+\small{HTL - ft (Ours)} & 0.81 \tiny{ $\pm$1e-03} & 0.53 \tiny{ $\pm$5e-03} & 0.47 \tiny{ $\pm$4e-03} & 0.64 \tiny{ $\pm$4e-03} \\
LOAN+\small{CTL - ft (Ours)} & 0.86 \tiny{ $\pm$7e-04} & 0.83 \tiny{ $\pm$5e-03} & 0.73 \tiny{ $\pm$4e-03} & 0.85 \tiny{ $\pm$3e-03} \\
 \cmidrule(r){1-5}
 LOAN+\small{LTL - Full} & 0.95 \tiny{ $\pm$4e-05} & 1.00 \tiny{ $\pm$7e-04} & 0.94 \tiny{ $\pm$7e-04} & 0.97 \tiny{ $\pm$4e-04} \\
 LOAN+\small{SCL - Full} & 0.97 \tiny{ $\pm$2e-05} & 1.00 \tiny{ $\pm$7e-04} & 0.97 \tiny{ $\pm$7e-04} & \textbf{0.98} \tiny{ $\pm$4e-04} \\
 LOAN+\small{CTL - Full (Ours)} & 0.96 \tiny{ $\pm$6e-05} & 0.99 \tiny{ $\pm$1e-03} & 0.95 \tiny{ $\pm$1e-03} & \textbf{0.98} \tiny{ $\pm$7e-04} \\
\bottomrule
\\
\multicolumn{5}{l}{\textbf{Wildfire}} \\
Model & \textit{Precision} & \textit{Accuracy} & \textit{IoU} & \textit{F1}\\
\midrule
FWI & 0.70 \tiny{ $\pm$0.002} & 0.57 \tiny{ $\pm$0.000} & 0.46 \tiny{ $\pm$0.00} & 0.63 \tiny{ $\pm$0.00} \\
\midrule
TimeSformer & 0.82 \tiny{ $\pm$7e-03} & 0.84 \tiny{ $\pm$0e+00} & 0.71 \tiny{ $\pm$5e-03} & 0.83 \tiny{ $\pm$4e-03} \\
 SwinTransformer3D & 0.72 \tiny{ $\pm$3e-03} & 0.89 \tiny{ $\pm$0e+00} & 0.66 \tiny{ $\pm$3e-03} & 0.80 \tiny{ $\pm$2e-03} \\
 LOAN \small{(Baseline)} & 0.92 \tiny{ $\pm$7e-03} & 0.85 \tiny{ $\pm$1e-16} & 0.79 \tiny{ $\pm$5e-03} & 0.88 \tiny{ $\pm$3e-03} \\
  \cmidrule(r){1-5}
 LOAN+\small{LTL - ft} & 0.97 \tiny{ $\pm$2e-03} & 0.98 \tiny{ $\pm$0e+00} & 0.95 \tiny{ $\pm$2e-03} & 0.97 \tiny{ $\pm$9e-04} \\
 LOAN+\small{SCL - ft} & 0.99 \tiny{ $\pm$2e-03} & 0.93 \tiny{ $\pm$1e-16} & 0.92 \tiny{ $\pm$1e-03} & 0.96 \tiny{ $\pm$7e-04} \\
 LOAN+\small{HTL - ft (Ours)} & 0.65 \tiny{ $\pm$2e-03} & 0.87 \tiny{ $\pm$0e+00} & 0.60 \tiny{ $\pm$2e-03} & 0.75 \tiny{ $\pm$2e-03} \\
LOAN+\small{CTL - ft (Ours)} & 0.84 \tiny{ $\pm$4e-03} & 0.87 \tiny{ $\pm$1e-16} & 0.74 \tiny{ $\pm$3e-03} & 0.85 \tiny{ $\pm$2e-03} \\
 \cmidrule(r){1-5}
 LOAN+\small{LTL - Full} & 1.00 \tiny{ $\pm$8e-04} & 0.94 \tiny{ $\pm$0e+00} & 0.94 \tiny{ $\pm$7e-04} & 0.97 \tiny{ $\pm$4e-04} \\
 LOAN+\small{SCL - Full} & 1.00 \tiny{ $\pm$7e-04} & 0.97 \tiny{ $\pm$0e+00} & 0.97 \tiny{ $\pm$7e-04} & \textbf{0.98} \tiny{ $\pm$4e-04} \\
 LOAN+\small{CTL - Full (Ours)} & 0.99 \tiny{ $\pm$2e-03} & 0.96 \tiny{ $\pm$0e+00} & 0.95 \tiny{ $\pm$1e-03} & \textbf{0.98} \tiny{ $\pm$7e-04} \\
\bottomrule
  \end{tabular}
  }
\label{tab:calabria_results_by_class}
\end{table}

\section{Margin Size: Ablation study}
\label{app:margin}

We conducted an ablation study to evaluate the effect of different margin values on classification performance, reported in Table~\ref{tab:app_margin_val}. 
Throughout the entire training phase for each model, the CL framework is employed.
The results indicate that increasing the margin size does not improve classification accuracy; on the contrary, in some cases, it may even adversely affect it. 
Based on the findings computed on the validation set, we choose the margin values for our approaches.

\begin{table*}[!ht]
\caption{Aggregated results computed over the year 2019 of the FireCube Dataset with different values for the margin. Each reported value represents the mean of five independent trials. We use this analysis to set the margin value in our experiments.}
  \centering
  \resizebox{.7\linewidth}{!}{
  \begin{tabular}{cl|c|c|c|c|c|c}
  \toprule
    PS & Model & \textit{Margin} & \textit{Precision} & \textit{Accuracy} & \textit{AUROC} & \textit{IoU} & \textit{F1}\\
    \midrule
\multirow{8}{*}{\rotatebox[origin=c]{90}{$1\times1$}}  & LOAN+\small{LTL - full} &5 & 0.85 \tiny{ $\pm$ 2e-02} & 0.85 \tiny{ $\pm$3e-02} & 0.93 \tiny{ $\pm$3e-03} & 0.74 \tiny{ $\pm$8e-03} & 0.85  \tiny{ $\pm$5e-03} \\
 & LOAN+\small{LTL - full} &10 & 0.87 \tiny{ $\pm$ 2e-02} & 0.87 \tiny{ $\pm$2e-02} & 0.93 \tiny{ $\pm$3e-03} & 0.77 \tiny{ $\pm$1e-02} & 0.87  \tiny{ $\pm$6e-03} \\
 & LOAN+\small{LTL - full} &20 & 0.88 \tiny{ $\pm$ 2e-02} & 0.88 \tiny{ $\pm$2e-02} & 0.95 \tiny{ $\pm$2e-03} & 0.78 \tiny{ $\pm$6e-03} & 0.88  \tiny{ $\pm$4e-03} \\
 & LOAN+\small{LTL - full} &50 & 0.87 \tiny{ $\pm$ 9e-03} & 0.87 \tiny{ $\pm$1e-02} & 0.94 \tiny{ $\pm$2e-03} & 0.77 \tiny{ $\pm$5e-03} & 0.87  \tiny{ $\pm$3e-03} \\
 \cmidrule(r){2-8}
 & LOAN+\small{CTL - full (Ours)} &5 & 0.88 \tiny{ $\pm$ 4e-03} & 0.88 \tiny{ $\pm$5e-03} & 0.94 \tiny{ $\pm$3e-03} & 0.78 \tiny{ $\pm$4e-03} & 0.88  \tiny{ $\pm$2e-03} \\
 & LOAN+\small{CTL - full (Ours)} &10 & 0.88 \tiny{ $\pm$ 4e-03} & 0.88 \tiny{ $\pm$5e-03} & 0.95 \tiny{ $\pm$3e-03} & 0.79 \tiny{ $\pm$5e-03} & 0.88  \tiny{ $\pm$3e-03} \\
 & LOAN+\small{CTL - full (Ours)} &20 & 0.89 \tiny{ $\pm$ 2e-02} & 0.89 \tiny{ $\pm$2e-02} & 0.94 \tiny{ $\pm$3e-03} & 0.80 \tiny{ $\pm$8e-03} & 0.89  \tiny{ $\pm$5e-03} \\
 & LOAN+\small{CTL - full (Ours)} &50 & 0.90 \tiny{ $\pm$ 1e-02} & 0.90 \tiny{ $\pm$1e-02} & 0.96 \tiny{ $\pm$2e-03} & 0.82 \tiny{ $\pm$5e-03} & 0.90  \tiny{ $\pm$3e-03} \\

\midrule
\multirow{8}{*}{\rotatebox[origin=c]{90}{$5\times5$}}  & LOAN+\small{LTL - full} &5 & 0.86 \tiny{ $\pm$ 4e-02} & 0.86 \tiny{ $\pm$5e-02} & 0.94 \tiny{ $\pm$3e-03} & 0.76 \tiny{ $\pm$1e-02} & 0.86  \tiny{ $\pm$8e-03} \\
 & LOAN+\small{LTL - full} &10 & 0.87 \tiny{ $\pm$ 5e-02} & 0.86 \tiny{ $\pm$7e-02} & 0.95 \tiny{ $\pm$3e-03} & 0.75 \tiny{ $\pm$2e-02} & 0.86  \tiny{ $\pm$1e-02} \\
 & LOAN+\small{LTL - full} &20 & 0.87 \tiny{ $\pm$ 4e-02} & 0.87 \tiny{ $\pm$5e-02} & 0.95 \tiny{ $\pm$3e-03} & 0.77 \tiny{ $\pm$1e-02} & 0.87  \tiny{ $\pm$8e-03} \\
 & LOAN+\small{LTL - full} &50 & 0.86 \tiny{ $\pm$ 5e-03} & 0.86 \tiny{ $\pm$6e-03} & 0.93 \tiny{ $\pm$4e-03} & 0.76 \tiny{ $\pm$6e-03} & 0.86  \tiny{ $\pm$4e-03} \\
 \cmidrule(r){2-8}
 & LOAN+\small{CTL - full (Ours)} &5 & 0.91 \tiny{ $\pm$ 2e-02} & 0.91 \tiny{ $\pm$2e-02} & 0.97 \tiny{ $\pm$1e-03} & 0.84 \tiny{ $\pm$4e-03} & 0.91  \tiny{ $\pm$2e-03} \\
 & LOAN+\small{CTL - full (Ours)} &10 & 0.91 \tiny{ $\pm$ 1e-02} & 0.91 \tiny{ $\pm$1e-02} & 0.97 \tiny{ $\pm$2e-03} & 0.83 \tiny{ $\pm$3e-03} & 0.91  \tiny{ $\pm$2e-03} \\
 & LOAN+\small{CTL - full (Ours)} &20 & 0.91 \tiny{ $\pm$ 2e-02} & 0.91 \tiny{ $\pm$2e-02} & 0.97 \tiny{ $\pm$1e-03} & 0.83 \tiny{ $\pm$6e-03} & 0.91  \tiny{ $\pm$4e-03} \\
 & LOAN+\small{CTL - full (Ours)} &50 & 0.89 \tiny{ $\pm$ 1e-02} & 0.89 \tiny{ $\pm$1e-02} & 0.96 \tiny{ $\pm$2e-03} & 0.81 \tiny{ $\pm$9e-03} & 0.89  \tiny{ $\pm$5e-03} \\
\midrule
\multirow{8}{*}{\rotatebox[origin=c]{90}{$15\times15$}} & LOAN+\small{LTL - full} &5 & 0.87 \tiny{ $\pm$ 4e-02} & 0.86 \tiny{ $\pm$6e-02} & 0.95 \tiny{ $\pm$2e-03} & 0.75 \tiny{ $\pm$1e-02} & 0.86  \tiny{ $\pm$9e-03} \\
 & LOAN+\small{LTL - full} &10 & 0.87 \tiny{ $\pm$ 5e-02} & 0.87 \tiny{ $\pm$6e-02} & 0.95 \tiny{ $\pm$3e-03} & 0.76 \tiny{ $\pm$1e-02} & 0.86  \tiny{ $\pm$9e-03} \\
 & LOAN+\small{LTL - full} &20 & 0.87 \tiny{ $\pm$ 4e-02} & 0.87 \tiny{ $\pm$6e-02} & 0.95 \tiny{ $\pm$3e-03} & 0.77 \tiny{ $\pm$1e-02} & 0.87  \tiny{ $\pm$9e-03} \\
 & LOAN+\small{LTL - full} &50 & 0.86 \tiny{ $\pm$ 9e-03} & 0.86 \tiny{ $\pm$1e-02} & 0.94 \tiny{ $\pm$4e-03} & 0.76 \tiny{ $\pm$7e-03} & 0.86  \tiny{ $\pm$5e-03} \\
 \cmidrule(r){2-8}
 & LOAN+\small{CTL - full (Ours)} &5 & 0.92 \tiny{ $\pm$ 1e-02} & 0.92 \tiny{ $\pm$1e-02} & 0.97 \tiny{ $\pm$1e-03} & 0.85 \tiny{ $\pm$7e-03} & 0.92  \tiny{ $\pm$4e-03} \\
 & LOAN+\small{CTL - full (Ours)} &10 & 0.89 \tiny{ $\pm$ 6e-02} & 0.88 \tiny{ $\pm$8e-02} & 0.97 \tiny{ $\pm$1e-03} & 0.78 \tiny{ $\pm$2e-02} & 0.88  \tiny{ $\pm$1e-02} \\
 & LOAN+\small{CTL - full (Ours)} &20 & 0.91 \tiny{ $\pm$ 8e-03} & 0.91 \tiny{ $\pm$1e-02} & 0.97 \tiny{ $\pm$1e-03} & 0.83 \tiny{ $\pm$5e-03} & 0.91  \tiny{ $\pm$3e-03} \\
 & LOAN+\small{CTL - full (Ours)} &50 & 0.89 \tiny{ $\pm$ 4e-02} & 0.89 \tiny{ $\pm$6e-02} & 0.97 \tiny{ $\pm$2e-03} & 0.79 \tiny{ $\pm$1e-02} & 0.89  \tiny{ $\pm$7e-03} \\
\midrule
\multirow{8}{*}{\rotatebox[origin=c]{90}{$25\times25$}} & LOAN+\small{LTL - full} &5 & 0.86 \tiny{ $\pm$ 2e-02} & 0.86 \tiny{ $\pm$3e-02} & 0.94 \tiny{ $\pm$3e-03} & 0.76 \tiny{ $\pm$9e-03} & 0.86  \tiny{ $\pm$6e-03} \\
 & LOAN+\small{LTL - full} &10 & 0.86 \tiny{ $\pm$ 5e-02} & 0.86 \tiny{ $\pm$7e-02} & 0.94 \tiny{ $\pm$3e-03} & 0.75 \tiny{ $\pm$2e-02} & 0.86  \tiny{ $\pm$1e-02} \\
 & LOAN+\small{LTL - full} &20 & 0.86 \tiny{ $\pm$ 6e-02} & 0.85 \tiny{ $\pm$9e-02} & 0.94 \tiny{ $\pm$2e-03} & 0.74 \tiny{ $\pm$2e-02} & 0.85  \tiny{ $\pm$1e-02} \\
 & LOAN+\small{LTL - full} &50 & 0.87 \tiny{ $\pm$ 3e-02} & 0.86 \tiny{ $\pm$5e-02} & 0.93 \tiny{ $\pm$3e-03} & 0.76 \tiny{ $\pm$1e-02} & 0.86  \tiny{ $\pm$7e-03} \\
 \cmidrule(r){2-8}
 & LOAN+\small{CTL - full (Ours)} &5 & 0.92 \tiny{ $\pm$ 1e-02} & 0.92 \tiny{ $\pm$1e-02} & 0.97 \tiny{ $\pm$1e-03} & 0.85 \tiny{ $\pm$5e-03} & 0.92  \tiny{ $\pm$3e-03} \\
 & LOAN+\small{CTL - full (Ours)} &10 & 0.92 \tiny{ $\pm$ 2e-02} & 0.92 \tiny{ $\pm$2e-02} & 0.98 \tiny{ $\pm$1e-03} & 0.85 \tiny{ $\pm$4e-03} & 0.92  \tiny{ $\pm$3e-03} \\
 & LOAN+\small{CTL - full (Ours)} &20 & 0.92 \tiny{ $\pm$ 2e-02} & 0.92 \tiny{ $\pm$3e-02} & 0.97 \tiny{ $\pm$2e-03} & 0.85 \tiny{ $\pm$6e-03} & 0.92  \tiny{ $\pm$4e-03} \\
 & LOAN+\small{CTL - full (Ours)} &50 & 0.91 \tiny{ $\pm$ 3e-02} & 0.91 \tiny{ $\pm$4e-02} & 0.97 \tiny{ $\pm$2e-03} & 0.83 \tiny{ $\pm$8e-03} & 0.91  \tiny{ $\pm$5e-03} \\

\bottomrule
  \end{tabular}
  }
\label{tab:app_margin_val}
\end{table*}

\end{document}